\documentclass[10pt, conference, compsocconf]{IEEEtran}
\IEEEoverridecommandlockouts
\usepackage{cite}
\usepackage{amsmath,amssymb,amsfonts}
\usepackage{algorithmic}
\usepackage{graphicx}
\usepackage{textcomp}
\usepackage{xcolor}

\usepackage[english]{babel}
\usepackage{booktabs} 
\usepackage[absolute,overlay]{textpos}
\usepackage{etoolbox}
\usepackage[figurename=Fig.,font={small}]{caption}
\usepackage[export]{adjustbox}
\usepackage[linesnumbered,ruled,vlined]{algorithm2e}
\usepackage{color, colortbl}
\usepackage{longtable}
\usepackage{xcolor}
\usepackage{soul}
\usepackage{comment}
\usepackage{url}
\usepackage{balance}

\usepackage{subfigure}
\usepackage{multirow}
\usepackage{lipsum}
\usepackage{tikz}
\usepackage{hyperref}
\usepackage{cleveref}
\usepackage{float}
\usepackage{xspace}
\usepackage{stfloats}

\soulregister\cite7 
\soulregister\citep7 
\soulregister\citet7 
\soulregister\ref7 
\soulregister\pageref7 
\soulregister\Cref7 

\newcommand{\sysname}{DiTMoS\xspace}

\begin{document}

\title{DiTMoS: Delving into Diverse Tiny-Model Selection on Microcontrollers\vspace{-0em}}
\author{\IEEEauthorblockN{Xiao Ma, Shengfeng He, Hezhe Qiao, Dong Ma$^{\ast}$ \thanks{*Dong Ma is the corresponding author.}}
\IEEEauthorblockA{\textit{School of Computing and Information Systems} \\ \textit{Singapore Management University} \\
\textit{Singapore} \\
\{xiaoma.2022, hezheqiao.2022\}@phdcs.smu.edu.sg, \{shengfenghe,dongma\}@smu.edu.sg}}


\maketitle

\begin{abstract}
\footnote{This paper has been accepted at PerCom 2024 and won the Best Paper Award.}
Enabling efficient and accurate deep neural network (DNN) inference on microcontrollers is non-trivial due to the constrained on-chip resources. Current methodologies primarily focus on compressing larger models yet at the expense of model accuracy. 
In this paper, we rethink the problem from the inverse perspective by constructing small/weak models directly and improving their accuracy.
Thus, we introduce \sysname, a novel DNN training and inference framework with a \textit{selector-classifiers} architecture, where the selector routes each input sample to the appropriate classifier for classification.
\sysname is grounded on a key insight: a composition of weak models can exhibit high diversity and the union of them can significantly boost the accuracy upper bound. 
To approach the upper bound, \sysname introduces three strategies including diverse training data splitting to increase the classifiers' diversity, adversarial selector-classifiers training to ensure synergistic interactions thereby maximizing their complementarity, and heterogeneous feature aggregation to improve the capacity of classifiers. 
We further propose a network slicing technique to alleviate the extra memory overhead incurred by feature aggregation. 
We deploy \sysname on the Neucleo STM32F767ZI board and evaluate it based on three time-series datasets for human activity recognition, keywords spotting, and emotion recognition, respectively. The experiment results manifest that: (a) \sysname achieves up to 13.4\% accuracy improvement compared to the best baseline; (b) network slicing almost completely eliminates the memory overhead incurred by feature aggregation with a marginal increase of latency. Code is released at {\href{https://github.com/TheMaXiao/DiTMoS}{https://github.com/TheMaXiao/DiTMoS}}

\end{abstract}



\section{Introduction}



Internet of Things (IoTs), equipped with diverse miniaturized and low-power sensors, have catalyzed a remarkable surge in continuous and cost-effective pervasive sensing applications in recent years, encompassing environmental monitoring~\cite{ullo2020iotenvironment}, asset tracking~\cite{khalid2022iotasset}, and on-body human sensing~\cite{anikwe2022iothealth}. On the other hand, advancements in artificial intelligence, especially deep neural networks (DNNs), enable effective extraction of both the explicit and implicit information within the substantial volume of sensor data, leading to enhanced sensing performance and more precise comprehension of the context~\cite{zhang2022deepsensingreview}. For example, recent transformer DNNs like SwinV2~\cite{liu2022swin} can boost the image recognition accuracy up to 90.17\% compared to 52.9\% achievable by conventional machining learning like support vector machine (SVM) on the ImageNet dataset \cite{lin2011large}.  

Nevertheless, realizing the potential of DNNs on IoTs for on-device pervasive sensing presents several challenges due to the inherent constraints of IoT devices. First, modern DNNs typically comprise millions of parameters and demand substantial memory for execution, whereas IoT devices are usually equipped with microcontroller units (MCUs) featuring constrained memory and storage. For example, mainstream MCUs normally come with SRAM of 16KB-256KB and Flash of 256KB-2MB)~\cite{stm32mainstream}, which are orders of magnitude smaller than what is required. Second, with such a memory budget, neural networks on MCUs have to be of smaller sizes, which results in diminished representation capability and consequently, poorer sensing performance\cite{banbury2021micronets}. Third, the limited computation capability and battery capacity of IoTs lead to prolonged inference time, which can be detrimental to latency-sensitive applications\cite{garbay2021cost}.

\begin{figure}[t]

  \subfigure[Top-down]{
		\includegraphics[width = 0.195\textwidth]{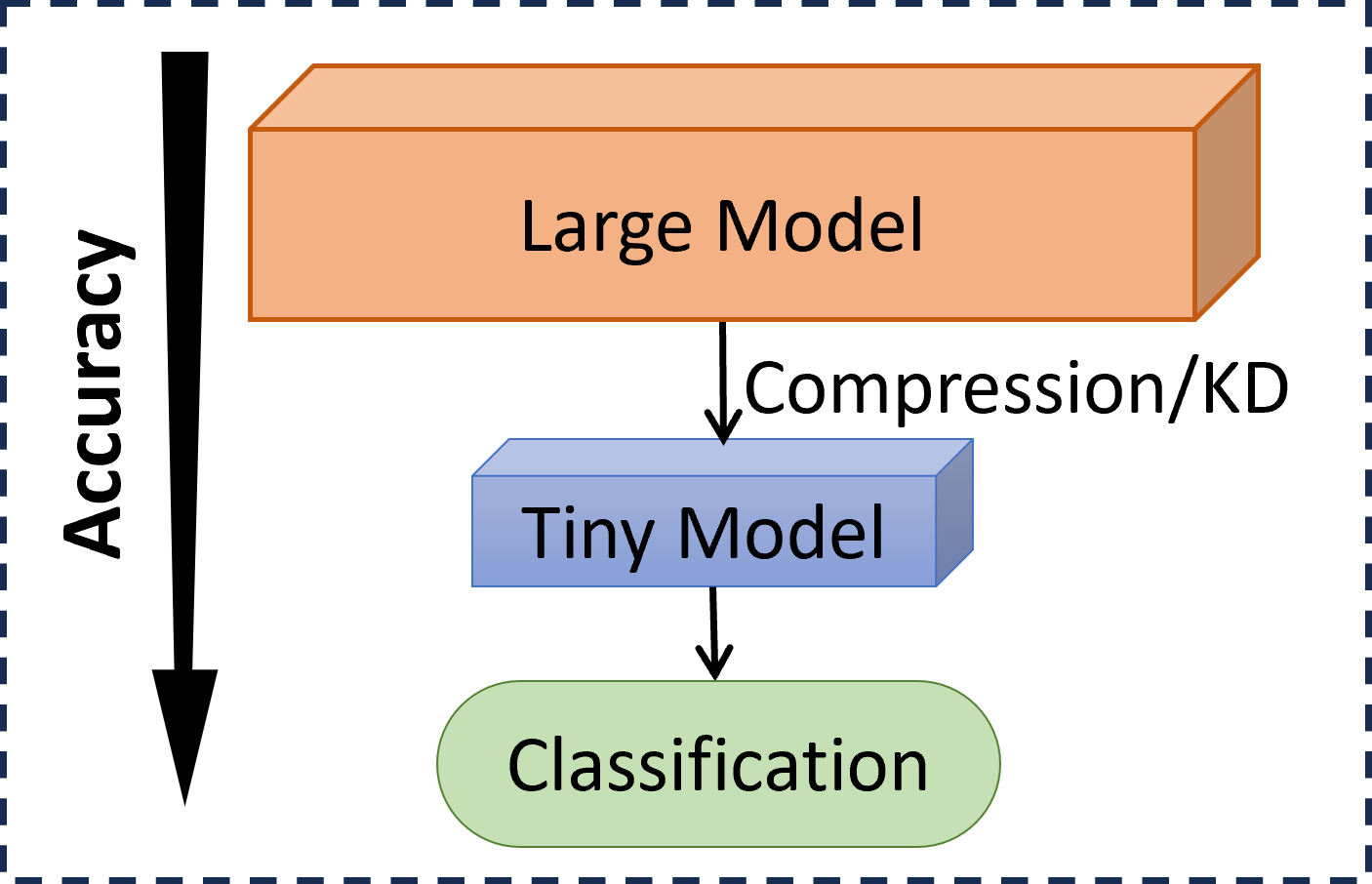}	
		\label{fig:topdown}} \subfigure[Bottom-up]{
		\includegraphics[width = 0.265\textwidth]{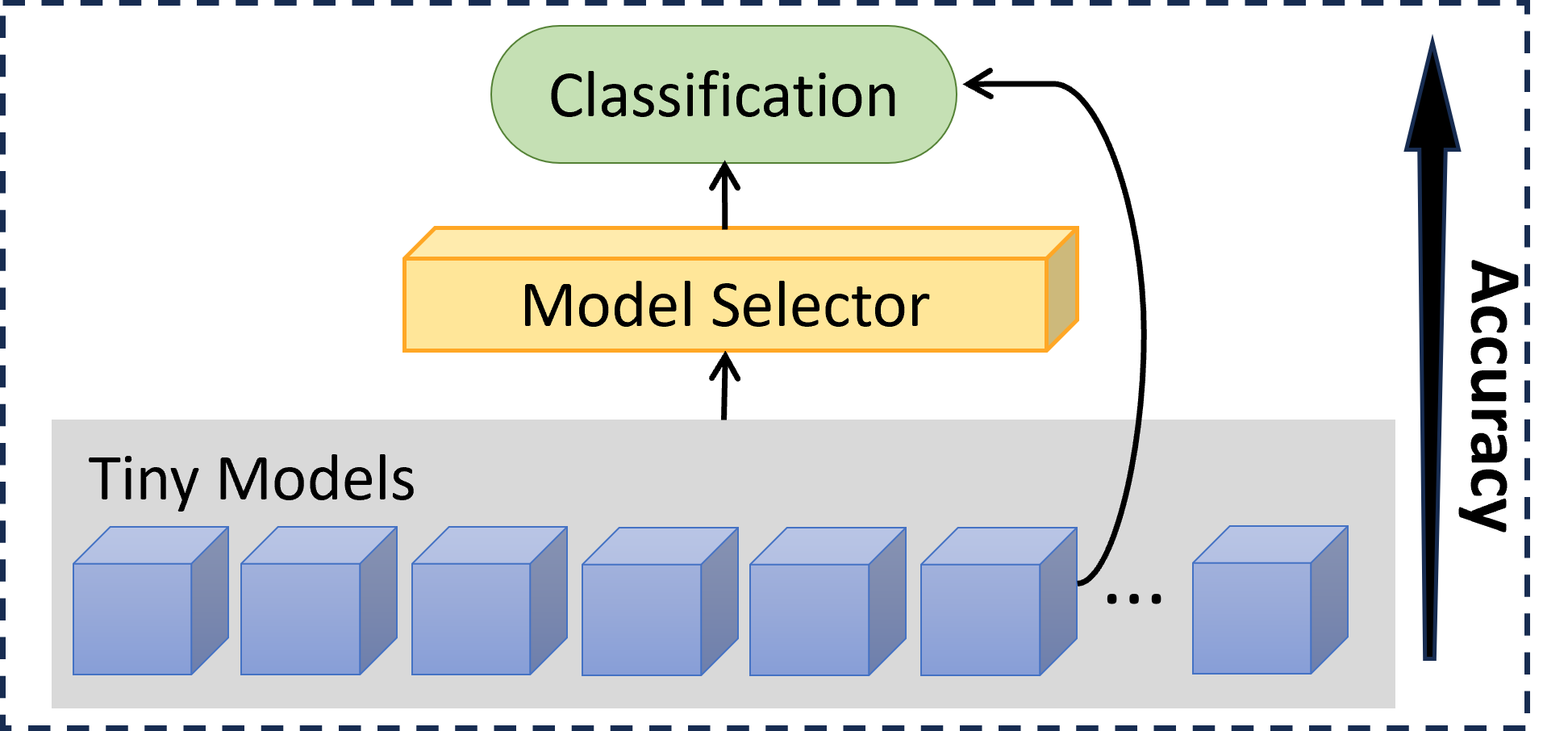}	
		\label{fig:bottmup}}
  \vspace{0in}
  \caption{Illustration of (a) top-down and (b) bottom-up methodology.}
  \label{fig: topbottom}
\vspace{-2mm}
\end{figure}

To address these challenges, current approaches employ a top-down methodology (\Cref{fig: topbottom}(a)), i.e., converting a large yet accurate DNN model into a smaller one with the objectives of minimizing accuracy loss, memory usage, and computation. Typical techniques include pruning which removes unnecessary or unimportant weights or connections \cite{liberis2023differentiable}, and quantization which reduces the precision of the weights and activations \cite{daghero2022human}. More recently, knowledge distillation (KD) has been exploited to distill the knowledge acquired by a strong model into a weaker one, facilitating the deployment on MCUs \cite{cerutti2020compact,brutti2022optimizing}. While these methods have proven effective, they adhere to a top-down principle that encounters a transferability limit when applied to tiny models. 

In this work, we rethink the problem from a bottom-up perspective by creating tiny models directly and elevating their accuracy to the next level, as shown in \Cref{fig: topbottom}(b). We have identified two key insights (\Cref{sec: model_diversity}): (1) smaller/weaker models can demonstrate greater diversity compared to their larger/stronger counterparts, owing to their inherent limited representation capability, and (2) the aggregation of multiple weak models promises a higher upper bound on classification accuracy (i.e., the percentage of samples correctly classified by \textit{at least} one classifier, refer to as \textit{union accuracy}). Building upon these insights, we propose \sysname, a hierarchical selector-classifiers architecture, where the selector extracts coarse-grained features from each input sample to determine the appropriate classifier for classification.

However, training an accurate and efficient selector-classifiers network encounters several challenges. First, model diversity plays a critical role in pushing the overall accuracy upper bound, while the limited capacity of weak models makes it challenging to control the training process when maximizing diversity across multiple models. Second, the overall accuracy is determined by both the selector's routing precision and the individual classifier's classification accuracy. Thus, there necessitates a close coupling between the selector and classifiers, which is challenging to achieve due to (1) the distinctive objectives of the two tasks and (2) the reliance on classifiers' diversity when training the selector. Third, compared to a pure-classifier structure (i.e., all the layers are used for classification), the selector network allocates a portion of the model parameters, resulting in a diminished classifier's capacity, and consequently, a reduction in classification accuracy. 

We tackle these challenges with three strategies: (1) we optimize classifiers' diversity by partitioning the training dataset into multiple subsets, based on prior semantic knowledge from a larger pre-trained model. Then, each classifier is initially trained with a separate subset to ensure they acquire knowledge from distinctive data sources; (2) we design an asynchronous and iterative adversarial training mechanism to enable synergistic interaction between the selector and classifiers, thereby maximizing their complementarity; (3) during classifiers training, we propose to re-utilize the feature extraction layers in the selector network (i.e., heterogeneous feature aggregation) to maximize the knowledge diversity employed in classifier classification. To mitigate the additional memory cost associated with heterogeneous feature aggregation, we propose a network slicing technique that reallocates the intermediate features from memory to the Flash, based on the memory management scheme of MCUs. 

We implement \sysname on the STM32F767ZI Nucleo board and assess its performance across three typical pervasive sensing applications that favor MCU deployment: human activity recognition (UniMiB-SHAR dataset), keyword spotting (Speech Commands dataset), and emotion recognition (DEAP dataset). The results reveal that \sysname consistently outperforms the best baselines, achieving an impressive accuracy enhancement of up to 13.4\%, while maintaining a similar level of memory usage and latency overhead. 

Our contributions are summarized as follows:
\begin{itemize}
    \item To the best of our knowledge, we are the first to uncover the high union accuracy of multiple weak models and propose a scheme to select the best one for classification.
    \item We present \sysname, a novel selector-classifiers framework designed for accurate and efficient DNN inference on MCUs. 
    \item We implement \sysname on real MCU platforms and showcase its superior performance over the state-of-the-art approaches under various configurations. 
\end{itemize}

\section{Preliminary}


\sysname leverages the model diversity to improve the classification performance. In this section, we first investigate the diversity of different models and introduce a naive model selection approach. 

\subsection{Model Diversity: Strong Model vs Weak Model}
\label{sec: model_diversity}
\begin{figure}[t]
 \centering
  \subfigure[Strong model]{
		\includegraphics[width = 0.255\textwidth]{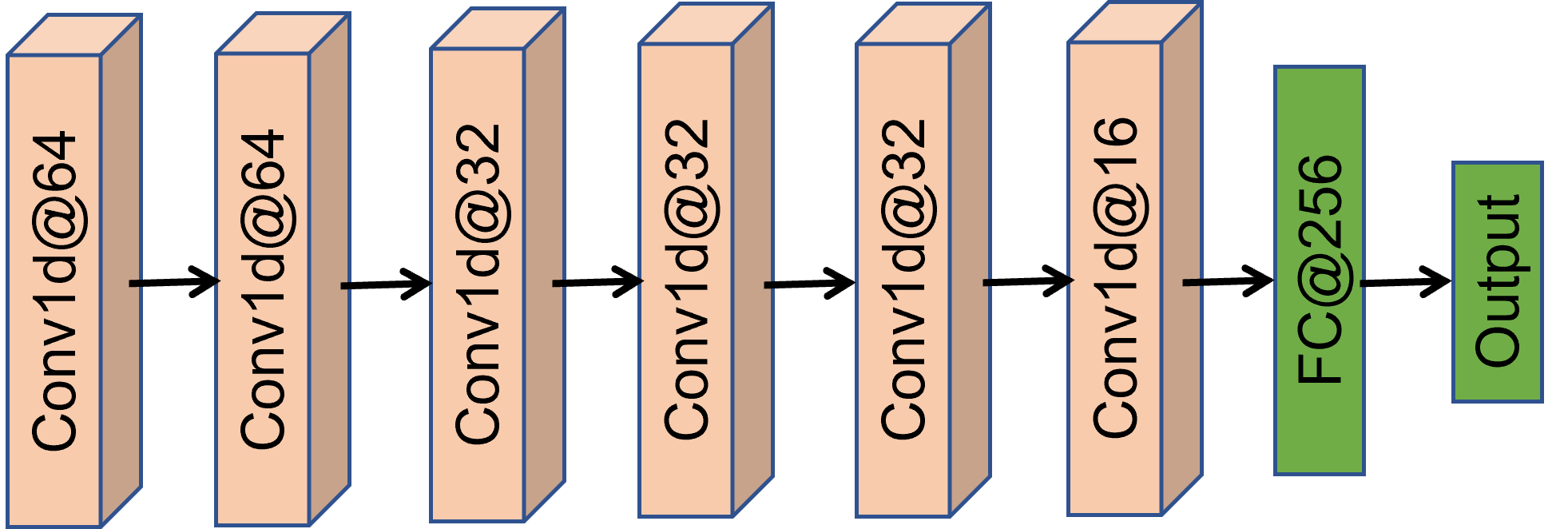}	
		\label{fig:architecture of strong model}}\hspace{4mm}\subfigure[Weak model]{
		\includegraphics[width = 0.145\textwidth]{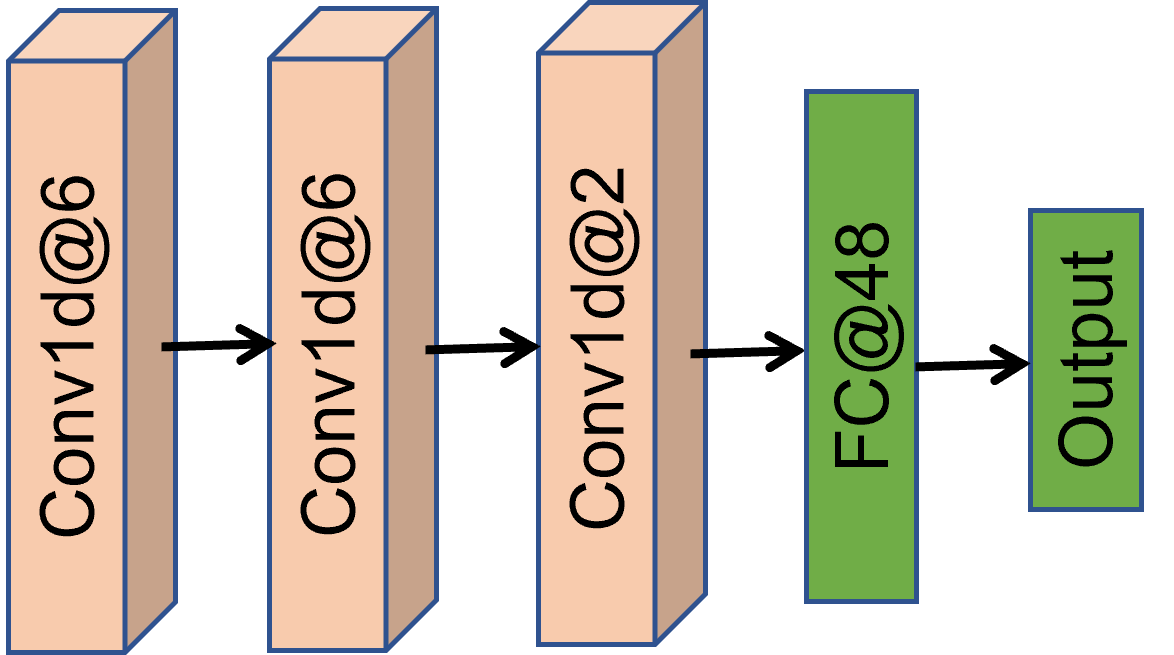}	
		\label{fig:architecture of weak model}}
\vspace{0in}
  \caption{The architecture of (a) strong model and (b) weak model.}
  \label{fig: model architecture}
   \vspace{-2mm}
\end{figure}

Deep learning models are trained in a data-driven manner, where the models automatically learn knowledge and skills from the data during the training process. How the model is trained (e.g., different hyperparameters, different input data, and different model initializations) will greatly influence the learning process\cite{yang2021taxonomizing}. Model diversity refers to the difference among the set of knowledge (i.e., feature representations) learned by different deep learning models, which is usually considered an effective property in improving the overall model accuracy. For example, ensemble learning combines a diverse set of models to boost the sensing performance\cite{huang2022deep}. 

However, we found that model diversity is closely associated with the capacity (i.e., the ability to represent and learn complex patterns from data) of a neural network. To demonstrate this, we conducted an experiment with a human activity recognition dataset named UniMiB-SHAR\cite{unimib}, which aims to classify 17 activities using a 3-axis accelerometer. We consider two levels of model capacity, i.e., strong and weak\footnote{Note that the `weak' and `strong' here refer to the classification performance on a given task, which is typically associated with the size (number of parameters) of the model, i.e., `small' and `large'.}. The strong model is composed of 6 one-dimensional convolutional layers (with a kernel size of 64, 64, 64, 32, 32, and 16, respectively) \footnote{Each convolutional layer is followed by a Max pooling layer in all the experiments.} and 1 fully-connected layer (with 256 neurons), while the weak model contains 3 one-dimensional convolutional layers (with a kernel size of 6, 6, and 2, respectively) and 1 fully-connected layer (with 48 neurons), as shown in \Cref{fig: model architecture}. Then, for each level, we trained five models with \textit{different initializations} (i.e., the model weights are initialized differently before training), during which all the hyperparameters remain the same.

\textbf{Centered Kernel Alignment (CKA)}: we first utilize the CKA, a metric that is widely used to measure the geometric similarity between two feature representations of a neural network\cite{kornblith2019similarity}, to quantify the model diversity. For two feature representations $X$ and $Y$, CKA can be computed as \cite{kornblith2019similarity} 
\begin{equation}
CKA(X,Y) = \frac{\| Y^{T}X \|_{F}^{2}}{ \| X^{T}X \|_{F} \| Y^{T}Y \|_{F}}
\end{equation}
where \( ||\cdot||_{F}\) is the Frobenius norm. CKA ranges from 0 to 1, where higher values reflect higher similarity (lower diversity). 

\begin{figure}[t]
\centering
  \subfigure[Strong models]{
		\includegraphics[width = 0.23\textwidth]{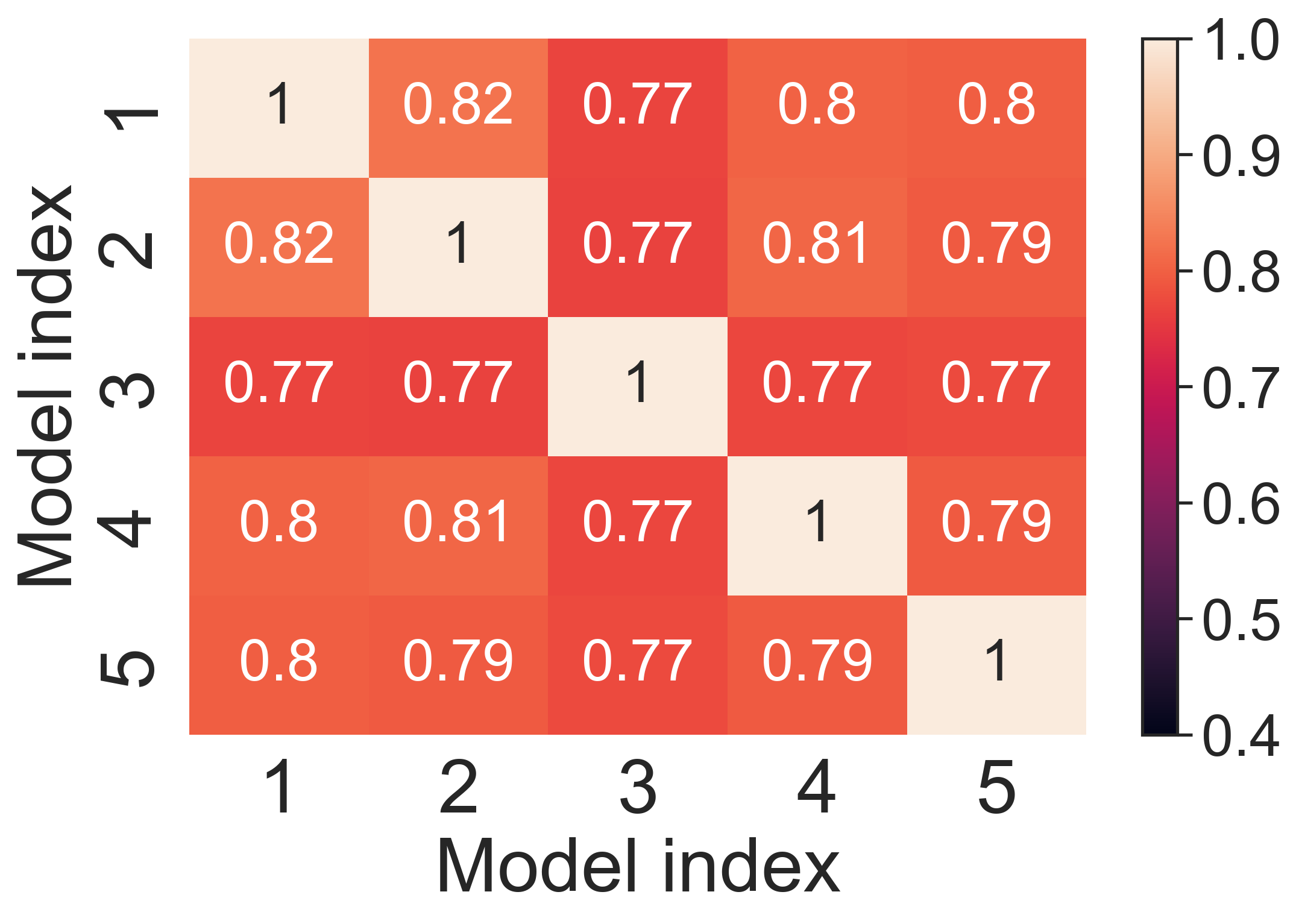}	
		\label{fig:CKA similarity of strong model group}}\hspace{0mm}\subfigure[Weak models]{
		\includegraphics[width = 0.23\textwidth]{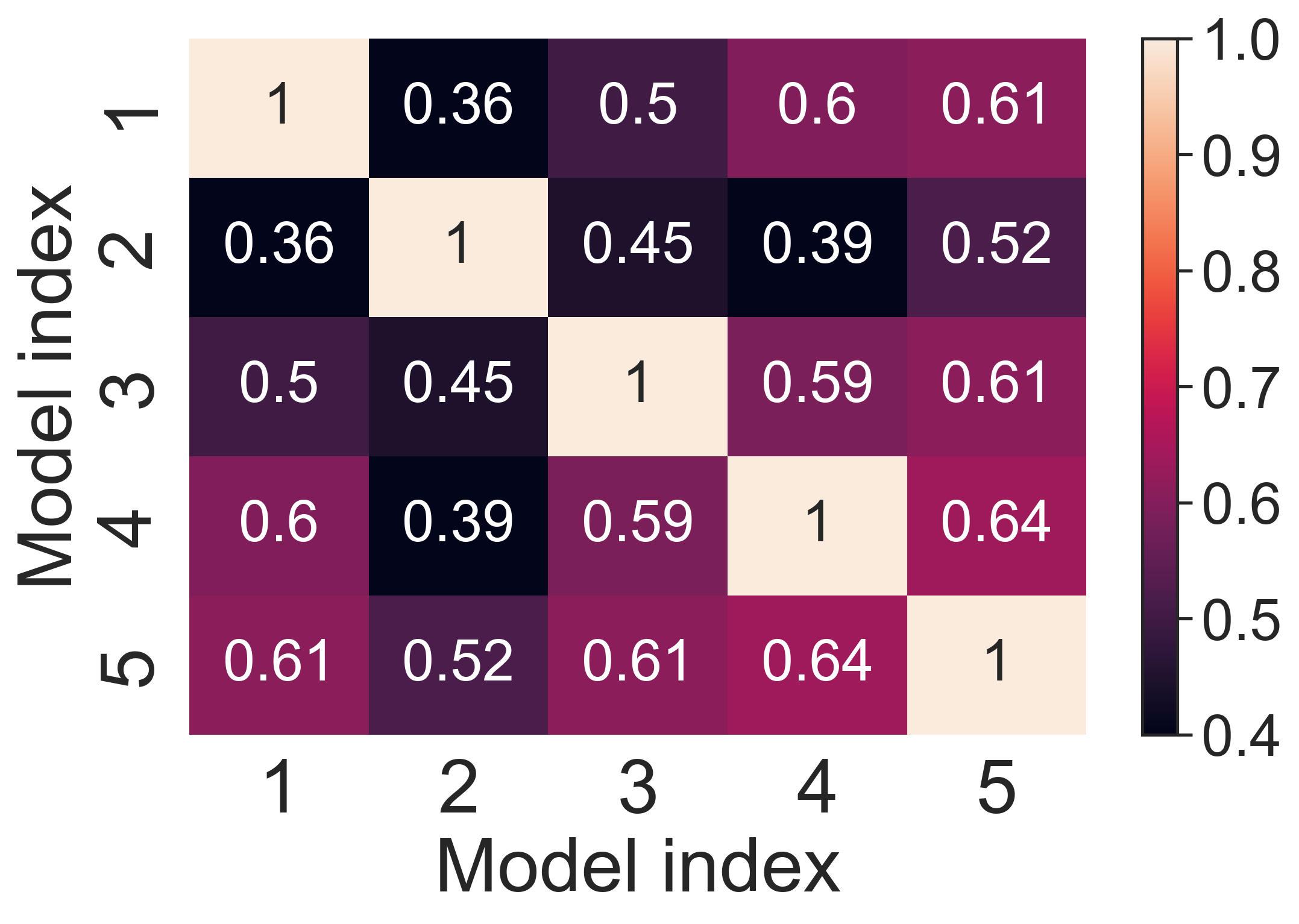}	
		\label{fig:CKA similarity of weak model group}}
 \vspace{0in}
 
  \caption{CKA matrix of (a) strong models and (b) weak models.}
  \label{fig: cka sim}
\vspace{-2mm}
\end{figure}

Specifically, we extracted the feature representations of the last convolutional layer and calculated the model-wise CKA among the five models, for each level. From \Cref{fig: cka sim}, we can observe that strong models yield higher CKA values than weak models, implying that models with higher capacity tend to converge to a similar set of features even though they are initialized differently. On the other hand, due to the limited capacity, weak models can only learn part of the knowledge and thus present higher diversity (lower CKA).   

\textbf{Classification Accuracy}: Then, we calculated the classification accuracy on the testing dataset and presented the results in \Cref{tab:tiny_large_performance}. All five strong models achieve high accuracy of around 95\% with a marginal difference, while the weak models obtain poorer accuracy with large variations. For example, the accuracy of Weak Model 1 is 6.9\% higher than that of Weak Model 2. This is because weak models learn more diverse features that have varying effectiveness on the same task. 

In addition, we further explore the diversity between models by defining an accuracy named \textit{union accuracy}, which represents the percentage of samples that can be correctly classified by \textbf{at least} one model. 
To calculate the union accuracy, we first performed inference with the five models and recorded the index of correctly classified samples. Then, we obtain a union set of the index by removing the repeated indexes. Finally, the union accuracy is computed as the ratio between the size of the union set and the total number of testing samples. As shown in the penultimate column of \Cref{tab:tiny_large_performance}, the union accuracy of the five strong models is only 3\% higher than the individual models, while weak models obtain a union accuracy that is around 20\% higher than the individual models. 

\begin{figure}[t]
\centering
\subfigure[Strong models]{
 \centering
 \includegraphics[width = 0.20\textwidth]{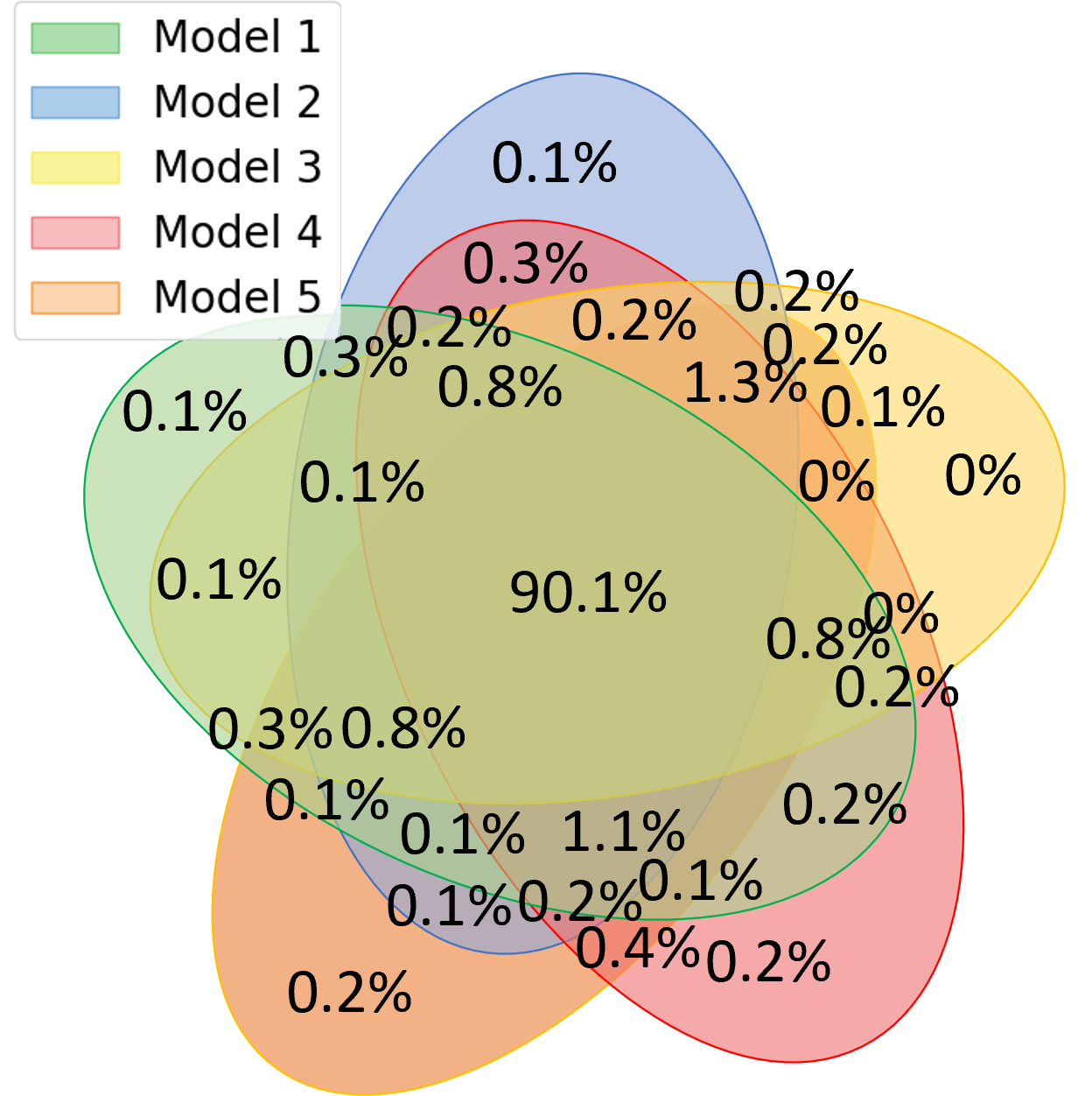}
  } \hspace{1mm}\subfigure[Weak models]{
  \centering
  \includegraphics[width = 0.20\textwidth]{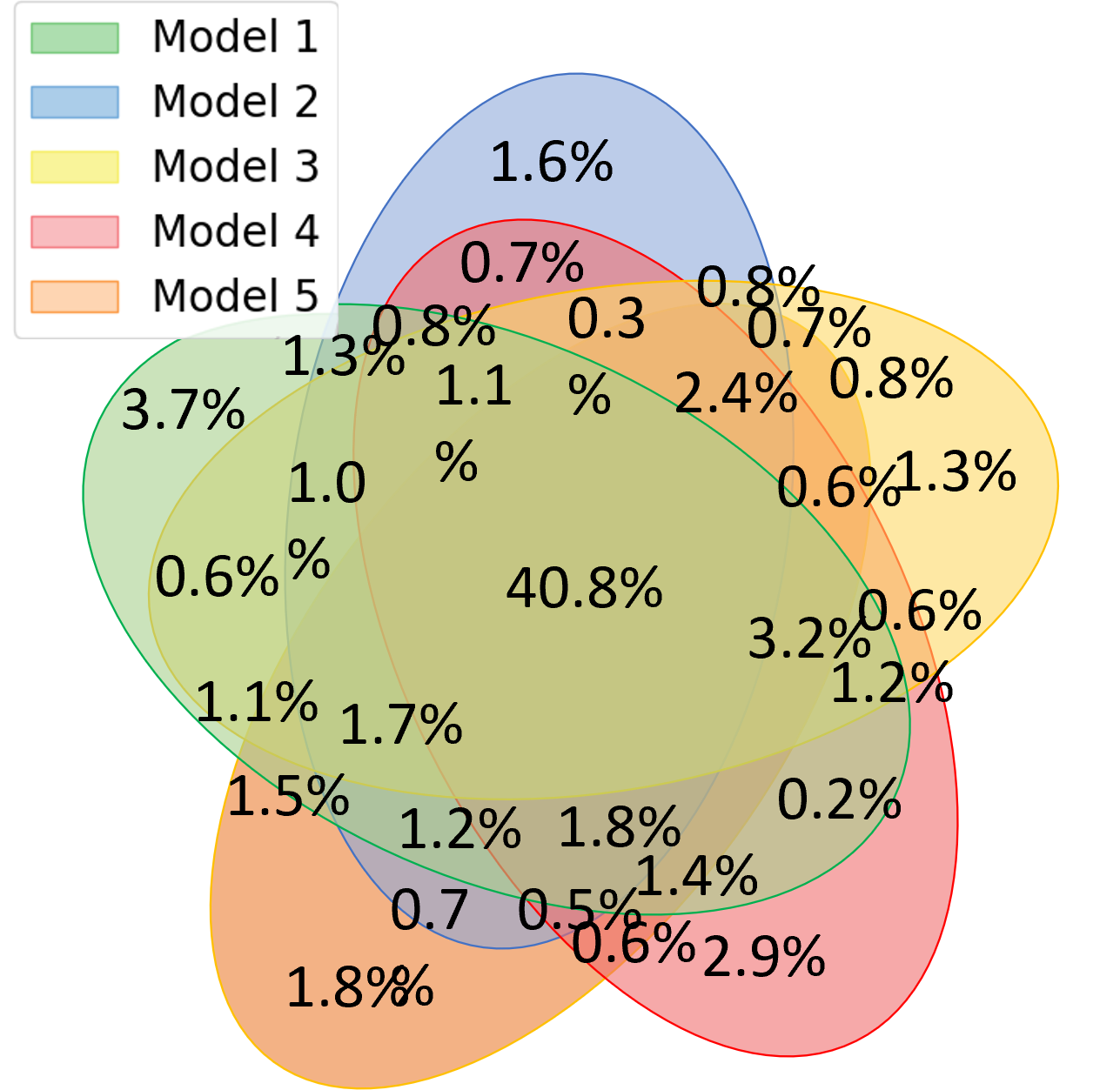}
  \label{fig:union_weak}
  }
  \vspace{0in}
  \caption{ Illustration of correctly classified sample index of the five models from the (a) strong group and (b) weak group.}
  \label{fig: union}
   \vspace{-2mm}
\end{figure}

\Cref{fig: union} further illustrates the detailed analysis of the classification results from the five strong models and five weak models respectively, where the numbers represent the percentage of samples being correctly classified. We can observe that (1) for each capacity level, most samples ( `easy' samples) can be correctly classified by all the individual models, i.e., overlapped knowledge learned by the models; (2) compared to strong models,  there are more samples that can \textbf{only} be correctly recognized by one weak model (i.e., higher model specialization) due to the diversity of weak models. The improved union accuracy offers a promise that \textbf{if we can accurately select the appropriate weak model for each input sample, the overall classification accuracy can be significantly enhanced.}


\begin{table}[t]
\centering
\caption{Comparison of model accuracy.}
\vspace{0mm}
\label{tab:tiny_large_performance}
\resizebox{0.99\linewidth}{!}{  
\begin{tabular}{cl|c|c|c|c|c|c|
>{\columncolor[HTML]{C0C0C0}}c}
\toprule
\multicolumn{2}{c|}{Model Index} & 1 & 2 & 3 & 4 & 5 & \textbf{\begin{tabular}[c]{@{}c@{}}Union\\ Accuracy\end{tabular}} & \textbf{\begin{tabular}[c]{@{}c@{}}Selection\\ Accuracy\end{tabular}} \\ \hline
\multicolumn{2}{c|}{Strong Model (484KB)} & 95.3\% & 95.9\% & 95.2\% & 96.1\% & 95.9\% & \textbf{98.9\%} & {\ul \textbf{}} \\ \hline
\multicolumn{2}{c|}{Weak Model (28KB)} & 64.7\% & 57.8\% & 58.2\% & 61.4\% & 60.9\% & \textbf{81.5\%} & \textbf{66.6\%} \\ \bottomrule
\end{tabular}%
}
 \vspace{0mm}
\end{table}

\subsection{A Naive Model Selection Approach}
\label{sec: simple_ms}

To realize this promise, an intuitive way is to train another deep learning model (referred to as \textit{selector}) to select the right weak model (referred to as \textit{classifier}) for each input. Thus, we first performed inference on the training dataset with the five weak models, which creates a new label for each training sample. For example, if a sample can only be correctly recognized by \textit{classifier 1} and \textit{classifier 3}, the new label is [1,0,1,0,0]. Then, we built a selector that has the same architecture as the weak models to avoid incurring additional computation overhead and trained it with the binary cross-entropy loss. The output of the selector is a vector indicating whether the corresponding classifier will be selected. 

The last column of \Cref{tab:tiny_large_performance} presents the performance of the above-mentioned model selection approach. We can see that the overall accuracy after model selection is only 66.6\% (merely 1.9\% higher than the best weak model), which is far away from the union accuracy. With a thorough analysis, we identified two main reasons that resulted in the poor performance of the above model selection. First, as shown in \Cref{fig:union_weak}, 40.8\% samples can be correctly classified by all models (i.e., labeled as [1,1,1,1,1]) and nearly 70.2\% samples are multi-labeled. \textit{Such a multi-label task is extremely challenging for a selector with the same architecture as the weak model}. Increasing the model capacity of the selector can improve the performance, but this also increases the computation overhead which is valuable on an MCU. Second,
the whole process can be considered a two-stage classification, where the selector aims to identify groups of samples with similar properties and the classifiers will be specialized on each group. So there would be some mutual correlations between the two stages that are useful for the overall classification. However, currently, \textit{the selector and classifiers are trained separately so that the mutual information cannot be captured}.

\section{DiTMoS Design}
\label{sec: system_design}

Based on the above findings, we propose \sysname, a novel deep learning framework that aims to improve the classification performance of tiny models on MCUs.
In this section, we first present an overview of the framework, followed by a detailed design of each block in the framework. 
\begin{figure*}[t]
\centering
  \includegraphics[width = \linewidth]{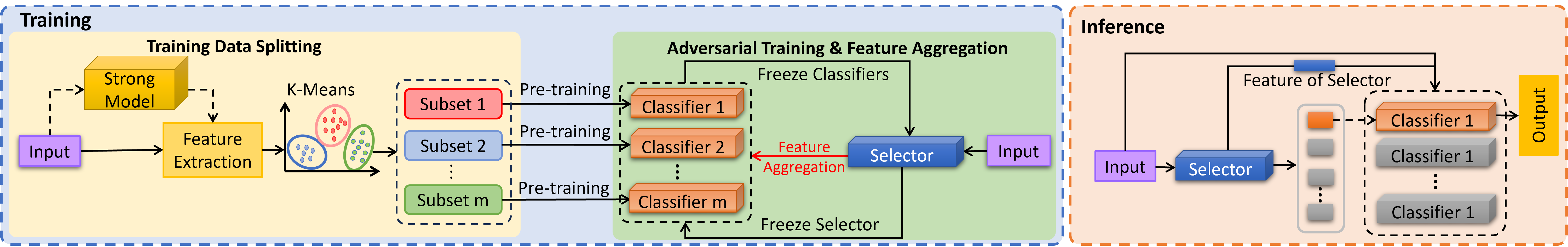}
  \caption{System overview of \sysname, including a training phase (left) and an inference phase (right).} 
  \label{fig: overview}
   \vspace{-2mm}
\end{figure*}

\subsection{Overview}
As illustrated in \Cref{fig: overview}, \sysname consists of a training phase on the server (GPU/CPU) and an inference phase on the MCU. The training phase leverages three main techniques: diverse training data splitting, selector-classifiers adversarial training, and heterogeneous feature aggregation. Specifically, to avoid the multi-label issue (i.e., one sample can be correctly classified by multiple classifiers), we propose to perform partitioning on the training dataset with prior semantic knowledge from a pre-trained strong model. Then, we train each classifier initially with the data from a particular subset, aiming to ensure that a sample can be correctly recognized by a classifier only. To allow the information coupling between the selector and classifiers, we design an adversarial training pipeline to train the selector and classifiers asynchronously and iteratively. In particular, 
one party (either the selector or classifiers) is frozen to allow the other to adapt during the training process. Meanwhile, considering the weak capability of classifiers, we propose to aggregate the heterogeneous feature representations from the selector when training the classifiers to improve classification performance. The training phase will output a trained selector and multiple trained classifiers. During inference, the input sample first undergoes the selector, which outputs the index of the selected classifier. Then, the raw input,  together with the intermediate feature representations of the selector, are fed into the selected classifier for classification.


\subsection{Diverse Training Data Splitting}


As mentioned in \Cref{sec: simple_ms}, classifiers trained with different initializations have a large overlap in terms of correctly classified samples. As a result, one sample can be routed to multiple classifiers when training the selector, which makes the training processing extremely challenging. The main reason is that each classifier is trained with the whole dataset, which allows the models to learn general yet similar knowledge especially when the model architecture is the same. So, to increase the specialization of classifiers, we should prevent a single classifier from seeing the whole training dataset. 

Thus, we propose to split the whole dataset into multiple subsets, and train a classifier on each subset. This design also speeds up the convergence of the training. Specifically, we first design a strong model, the same 6-layer CNN strong model as used in the \Cref{sec: model_diversity}, that can achieve nearly optimal performance on the classification task. Then, for each input, we extract the embeddings from the last convolutional layer of the pre-trained strong model as the corresponding features (as shown with the dotted line in \Cref{fig: overview}). {The motivation behind feature extraction is to enhance the separability of subsets and mitigate overlap among them. Given the high performance of the strong model, the extracted features exhibit a strong representation capability when splitting the dataset. Afterward, we apply the K-Means\footnote{Note that we tested various clustering algorithms including K-Means, K-Means++, GMM, and Mean Shift. The results indicate that the selection of the clustering algorithm has minimal impact on the final performance because the subsets will be refined by the following iterative training process.} \cite{macqueen1967some} algorithm on the extracted features to obtain multiple subsets, where the number of subsets $m$ is determined by the number of classifiers. Finally, each classifier is pre-trained with the corresponding subset to gain its specialization. \Cref{fig: clustering_effect} presents the detailed testing performance of the five weak models after splitting the training dataset of UniMiB-SHAR. We can observe that (1) although each individual classifier only achieves approximately 17\% accuracy on the full testing dataset, the union accuracy can be up to 82.1\%, implying the specialization of each classifier; (2) the overlap between different classifiers is only 3.6\%, demonstrating the effectiveness of training data splitting in alleviating the multi-label issue.


\begin{figure}[t]
\vspace{-12mm}
\centering
\begin{minipage}[b]{0.15\textwidth}
\centering
\begin{table}[H]
    \centering
    \vspace{1cm}
    \begin{tabular}{c|c}
    \toprule
    Model Index & Accuracy \\ \hline
    \multicolumn{1}{c|}{1} & \multicolumn{1}{c}{17.5\%} \\ \hline
    \multicolumn{1}{c|}{2} & \multicolumn{1}{c}{17.2\%} \\ \hline
    \multicolumn{1}{c|}{3} & \multicolumn{1}{c}{17.1\%} \\ \hline
    \multicolumn{1}{c|}{4} & \multicolumn{1}{c}{16.4\%} \\ \hline
    \multicolumn{1}{c|}{5} & \multicolumn{1}{c}{17.6\%} \\ \hline
    Union & 82.1\% \\ \bottomrule
    \end{tabular}%
    \vfill
    \caption*{(a)}
\end{table}
\end{minipage} \hspace{6mm} \begin{minipage}[b]{0.23\textwidth}
\centering
\begin{figure}[H]
    \centering
    \includegraphics[scale=0.28]{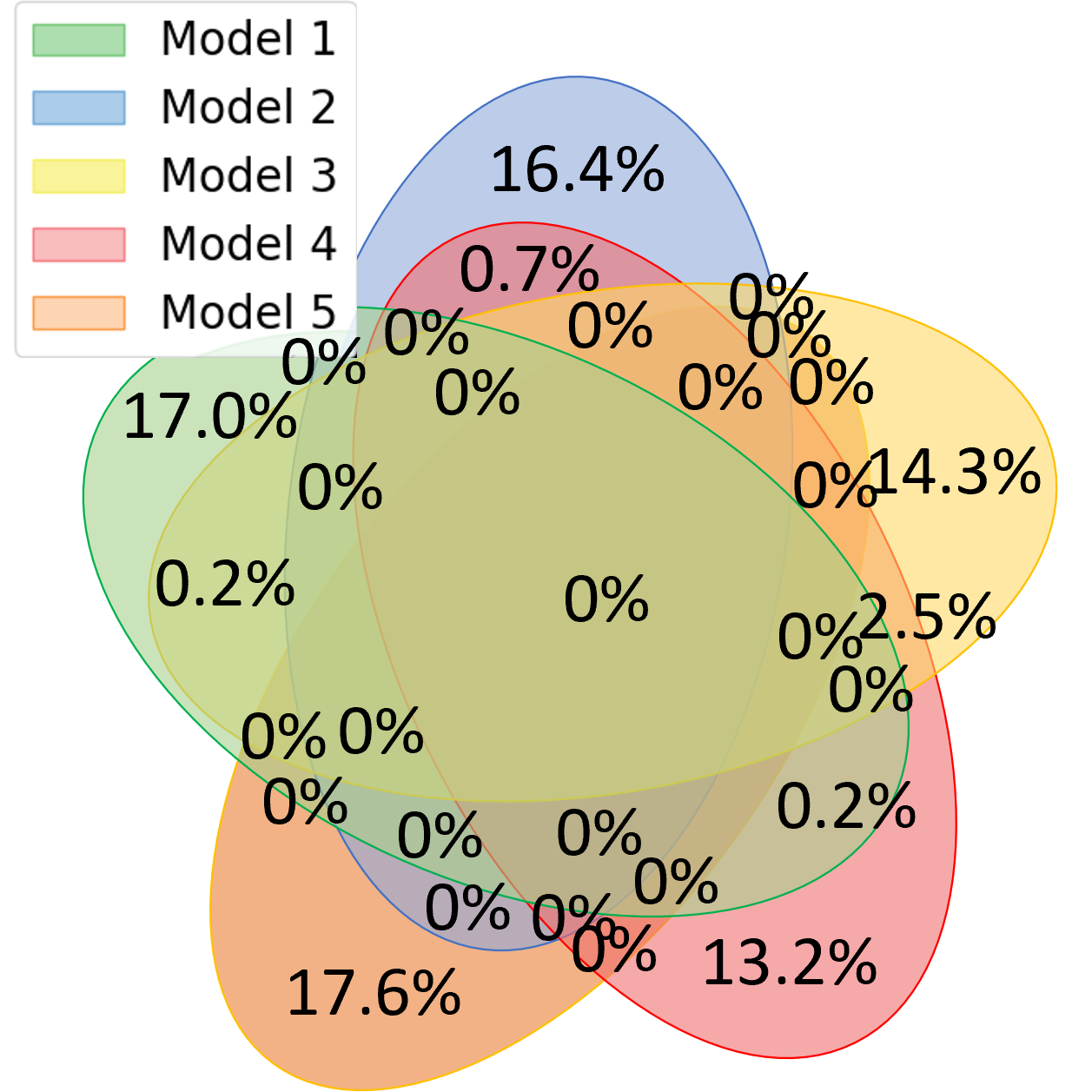}
    \caption*{(b)}
\end{figure}
\end{minipage}
 \vspace{0mm}
  \caption{ (a) Classification accuracies and (b) illustration of correctly classified sample index of the five models with training data splitting on the UniMiB dataset.}
  \label{fig: clustering_effect}
 \vspace{-2mm}
\end{figure}

\subsection{Adversarial Selector-Classifiers Training}
The selector and classifiers are coupled with each other. Specifically, to choose the best classifier for the current input sample, the selector should have some knowledge about the classifiers (i.e., understand the characteristics of different options). On the other hand, the classifiers should enlarge and associate their diversity with the selector to reduce the difficulty of selection.  
However, another problem identified from the experiment in \Cref{sec: simple_ms} is that training the selector and classifier independently ignores their mutual correlation. 

To gain the mutual correlation, a straightforward solution is to train the selector and classifiers synchronously. However, we experimentally found that this incurs two issues and leads to poor performance. First, unlike classifiers that have already been pre-trained on different subsets, the selector is randomly initialized and has no knowledge about the task. Such a knowledge gap would deteriorate the specializations (prior knowledge) of classifiers, making training data splitting ineffective. Second, in essence, the selector is responsible for model selection, while the classifiers are dedicated to fine-grained classification. Simultaneously training two models with distinct objectives presents challenges because they interact within different task-specific feature spaces. 



To address the problems,
we propose to train the selector and classifiers asynchronously, iteratively,  and in an adversarial manner. Specifically, one adversarial training iteration consists of two steps: (1) \textit{selector training}:  based on the classifiers' prior knowledge, the selector will be trained for several epochs (e.g., 5), during which the weights of the classifiers are frozen; (2) \textit{classifiers training}: after the selector learns the characteristics of the classifiers, the classifiers will be re-trained (fine-tuned) for several epochs, during which the selector weights are frozen. The two-step adversarial training will be executed for multiple iterations until convergence. Next, we introduce the two steps in detail.  


\textit{Selector training: } to train the selector, we first need to generate a label for each sample, i.e., the index of the classifier that can correctly classify it. We achieve this by performing inference on the training dataset using all the $m$ classifiers. However, the label generation process presents two issues. First, although the training data splitting is effective in reducing the knowledge overlap among the classifiers, it cannot completely eliminate the multi-label issue as some easy samples (around 3.6\% in our experiment) can still be recognized by different classifiers. For these multi-label samples, we convert their label to single-label by assigning the classifier with the highest confidence. Second, due to the weak capability of the classifiers, there are still some samples that cannot be correctly classified by any classifier, i.e., the one-hot label is [0, 0, ..., 0]. We also convert them into single-label samples by assigning the classifier with the lowest confidence. The underlying rationale is that the low-confidence classifier has a higher chance to correctly classify the sample by adjusting its weights during training, compared to high-confidence classifiers. After the labels are processed, the selector's weights will be trained to couple with the classifiers trained in the previous iteration.

\textit{Classifiers training: } unlike conventional single-classifier training, there are multiple classifiers that need to be trained individually on different input samples (i.e., each classifier can only perceive part of the training set). Based on the assignment of the selector, for each batch, only the weights of selected classifiers will be updated. For instance, if a batch contains 16 samples, in which 6 samples are routed to \textit{classifier 1}, 5 samples are routed to \textit{classifier 2}, and 5 samples are routed to \textit{classifier 3}, only the three classifiers will be trained and the rest $m-3$ classifiers will remain unchanged. Since the sample assignment is based on the selector (new subsets will be generated at each epoch), the samples fed to each classifier might be different from the subsets obtained during training data splitting in the beginning. However, this is preferable as (1) the initial training data splitting may not be optimal and (2) the adversarial training process actually fine-tunes the alignment between selector and classifiers over iterations.  

\textit{Loss function: }
although we freeze the classifiers when training the selector, and vice versa, our loss function needs to consider both stages as the aim is to align them iteratively. Specifically, as shown in \Cref{eq2}, we designed a loss function that consists of four components. In detail, $CE_{sel}$ represents the cross-entropy loss of the selector, which measures the performance of the selector in terms of correctly routing an input sample to the right classifier. The other three losses (i.e., $CE_{single}$, $CE_{union}$, $CE_{overlap}$) are all designed for the classifiers. Concretely, when there are multiple classifiers, the classification result of a sample can be in three cases: (i) recognized by a single classifier only, (ii) recognized by multiple classifiers simultaneously, and (iii) none of the classifiers can recognize it. Then, $CE_{single}$ is the cross-entropy loss on the samples in case (i), which only updates the corresponding classifier; $CE_{union}$ is the cross-entropy loss designed to deal with case (iii). It aims to improve the union accuracy by updating all classifiers when none of them can recognize a sample; $CE_{overlap}$ is the negative cross-entropy loss designed for case (ii), which aims to further increase the diversity of classifiers by reducing their knowledge overlap. All the three losses are sample-wise, meaning that different losses are applied to different samples, instead of the whole batch. We further introduce three coefficients, i.e., $\alpha$, $\beta$, and $\gamma$, to regularize the weights of three losses related to classifiers. The coefficients will impact the accuracy of DiTMoS. In our experiments, we conducted a grid search for the three coefficients with values ranging from 0.01 to 1 and the results reveal a 3\% difference between the least and most effective settings. Following the grid search results, we assign 0.1, 0.1, and 0.03 to $\alpha$, $\beta$, and $\gamma$ respectively that yield optimal performance. 
\vspace{0mm}
\begin{equation}
Loss =  CE_{sel} + \alpha\cdot CE_{single} + \beta\cdot CE_{union}  + \gamma\cdot CE_{overlap} \label{eq2}
\vspace{0mm}
\end{equation}

\subsection{Heterogeneous Feature Aggregation}
\label{sec: fea_reuse}


The subset-based classifier pre-training and adversarial selector-classifiers training result in diversified classifiers and the corresponding correlated selector. However, the selector-classifiers architecture has two inherent limitations that hinder it from achieving optimal performance. First, it splits the whole network into two parts, each focusing on sample routing and classification, respectively. Compared to classification using the full network, the classifier's representation capability is reduced, leading to diminished performance. Second, due to the routing of the selector, classifiers are trained on a subset of the data only, preventing them from observing the complete dataset. This limitation results in the loss of global and general features that could be advantageous for classification.

\begin{figure}[t]
\centering
  \includegraphics[width = 0.48\textwidth]{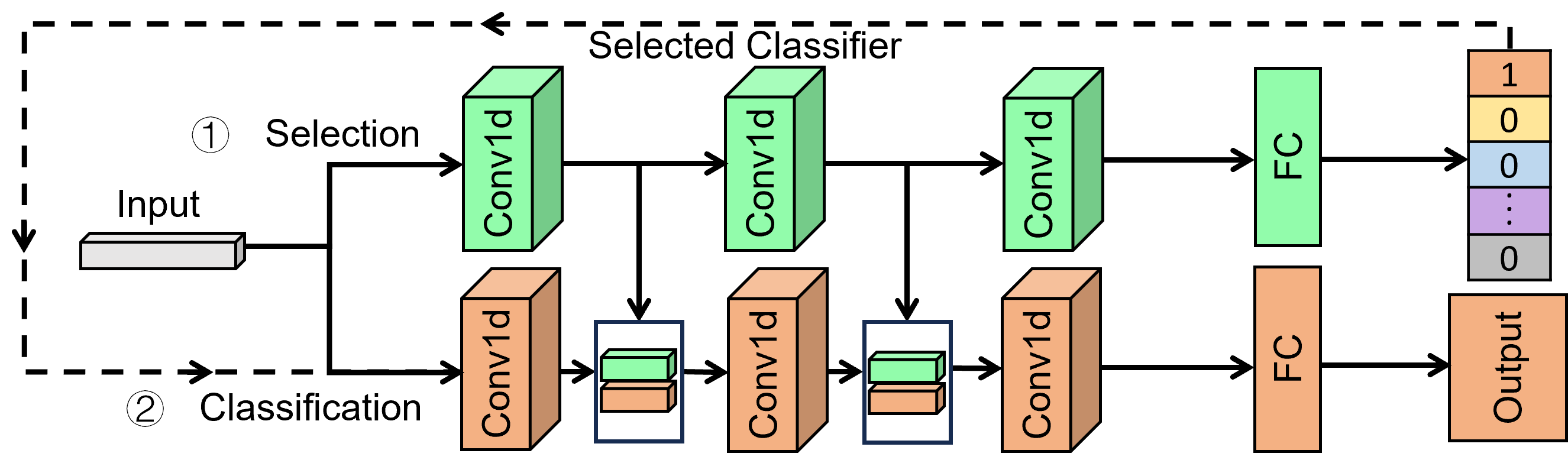}
  \caption{Illustration of heterogeneous feature aggregation.}
  \label{fig: feature reuse}
\vspace{-2mm}
\end{figure}

To overcome the two limitations and further improve the classification performance, we propose to aggregate the heterogeneous intermediate features from the selector when training the classifiers. Specifically, \Cref{fig: feature reuse} illustrates the overall structure of the network, which consists of three consecutive convolutional layers and a fully connected layer for the selector and classifiers, respectively\footnote{Note that we consider the same architecture for both the selector and classifiers in our evaluation to demonstrate the concept. However, their architectures can be designed arbitrarily based on the application requirements.}. An input sample first undergoes the selector for classifier selection and then is passed to the selected classifier for classification. During classification, the intermediate feature representations from the first and second convolutional layers of the selector are concatenated to the outputs of the corresponding layers in the selected classifier before feeding to the next layer, namely, heterogeneous feature aggregation. Such a strategy not only incorporates the heterogeneous global information from the full dataset but also increases the representation capability of the classifier by borrowing more convolutional layers from the selector for feature extraction. However, not all the borrowed features are helpful as deeper convolutional layers tend to generate features that are more specific to the corresponding task. We choose to aggregate the feature representations from the first two convolutional layers of the selector only because they exhibit a higher degree of task-neutrality. In contrast, we experimentally found that incorporating the third convolutional layer indeed leads to a deterioration in classification accuracy.

\subsection{Training Settings}
Our framework is implemented with PyTorch and trained on an NVIDIA 3090 GPU. In each adversarial training iteration, the selector and classifiers are trained for 6 epochs respectively. We utilize the SGD optimizer and set the initial learning rate as 0.001 with a decay of 0.5 after every 5 iterations.



\section{MCU Implementation} \label{sec: implemtation}


After training the selector and classifiers on the server, we deploy them on the MCU. Next, we describe the process of the deployment and how we optimize the memory usage.

\subsection{Platform and Workflow}
In this paper, we choose Nucleo STM32F767ZI~\cite{stm32f767zi} as the MCU test platform given its wide adoption in industrial IoTs\cite{banbury2021micronets,liberis2022pex}. It is equipped with a Cortex-M7 MCU, a 512~KB SRAM, and a 2~MB Flash. STM32 provides an official integrated development toolchain including STM32CubeIDE and its associated artificial intelligence (AI) package
for onboard inference of deep learning models. 

To deploy \sysname on the STM32 board, we first convert the PyTorch models for the selector and classifiers to the ONNX format (an open format for machine learning models that allows model interchange between different frameworks~\cite{onnx}). Then, the ONNX models are uploaded to the AI package to analyze the model's memory usage by loading and executing the model layer-by-layer. Specifically, only the parameters (trainable weights and constants) of the currently executed layer will be loaded from the Flash to the memory for computation. Meanwhile, all the intermediate tensors (i.e., activations) produced at runtime must be stored in the memory, constraining the peak memory usage~\cite{liberis2019reorder}. Finally, the toolchain generates the C-code to execute models.

\subsection{Memory Optimization with Network Slicing}

\begin{figure}[t]
\centering
  \includegraphics[width = 0.48\textwidth]{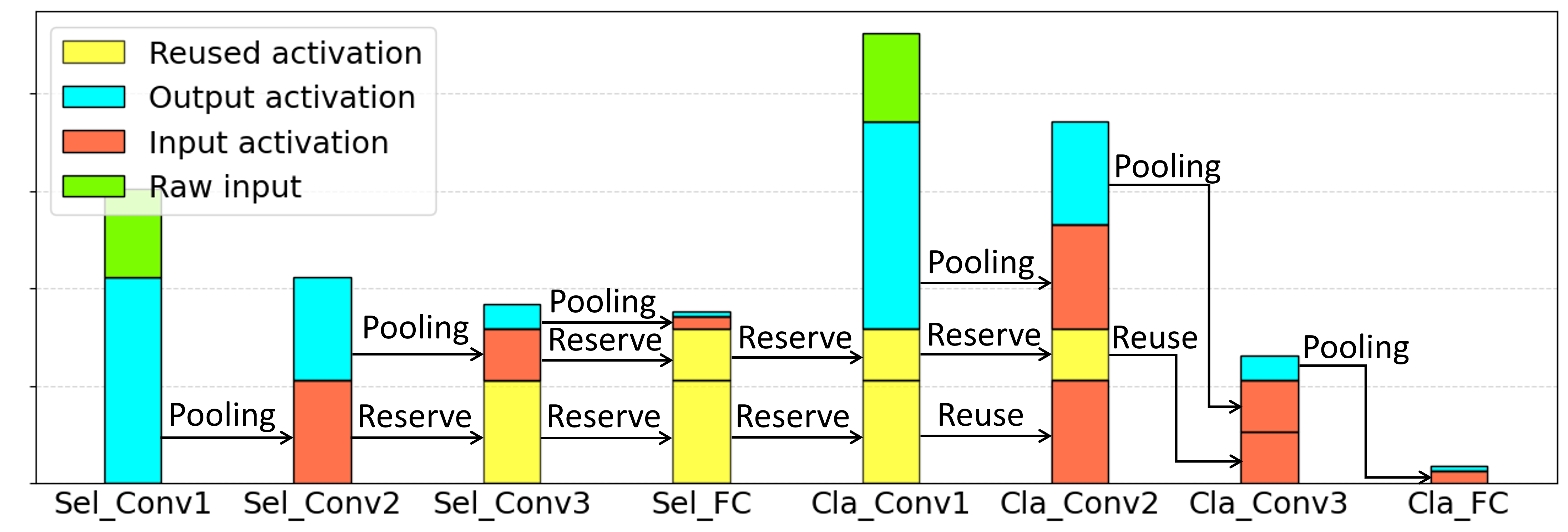}
  \vspace{0mm}
  \caption{Memory usage of \sysname without network slicing. }
  \label{fig: memory consumption}
   \vspace{-2mm}
\end{figure}
As presented in \ref{sec: fea_reuse}, \sysname aggregates the intermediate feature representations from the selector to improve the classification performance. Since the intermediate activations have to be held in the memory until they are integrated into the classifier, the peak memory of \sysname is increased. 
For example, \Cref{fig: memory consumption} illustrates the detailed memory usage of \sysname throughout the inference. 
Concretely, the outputs of the first convolutional layer of the selector will be retained in the memory after Maxpooling until it is reused in the first convolutional layer of the selected classifier, and similarly to the second convolutional layer of the selector. The peak memory usage is observed in the first convolutional layer of the classifier, with a significant portion occupied by reserved intermediate activations.

To eliminate the extra memory consumption incurred by feature aggregation, we propose to split the whole network into multiple slices, where each slice manages its input/output by directly communicating with the Flash. Specifically, as shown in \Cref{fig: network slicing}, both the selector and classifier are split into three slices, in which the first two convolutional layers are treated as independent slices respectively, and the third convolutional layer plus the fully connected layer are combined in separate slices. As a result, the output of slice 1 and slice 2 can be stored on the Flash and reloaded when they are used in slice 4 and slice 5. Such a slicing strategy is designed based on the characteristics of feature aggregation, and it almost completely removes the extra memory usage by migrating the to-be-reused intermediate activations to the Flash. However, the latency of the inference will be slightly increased because (1) storing and reloading the activations require extra time and (2) computing the aggregated features consumes more time. We will evaluate the overhead of network slicing in Section~\ref{sec: system_performance}. 

\begin{figure}[t]
\centering
  \includegraphics[width = 0.48\textwidth]{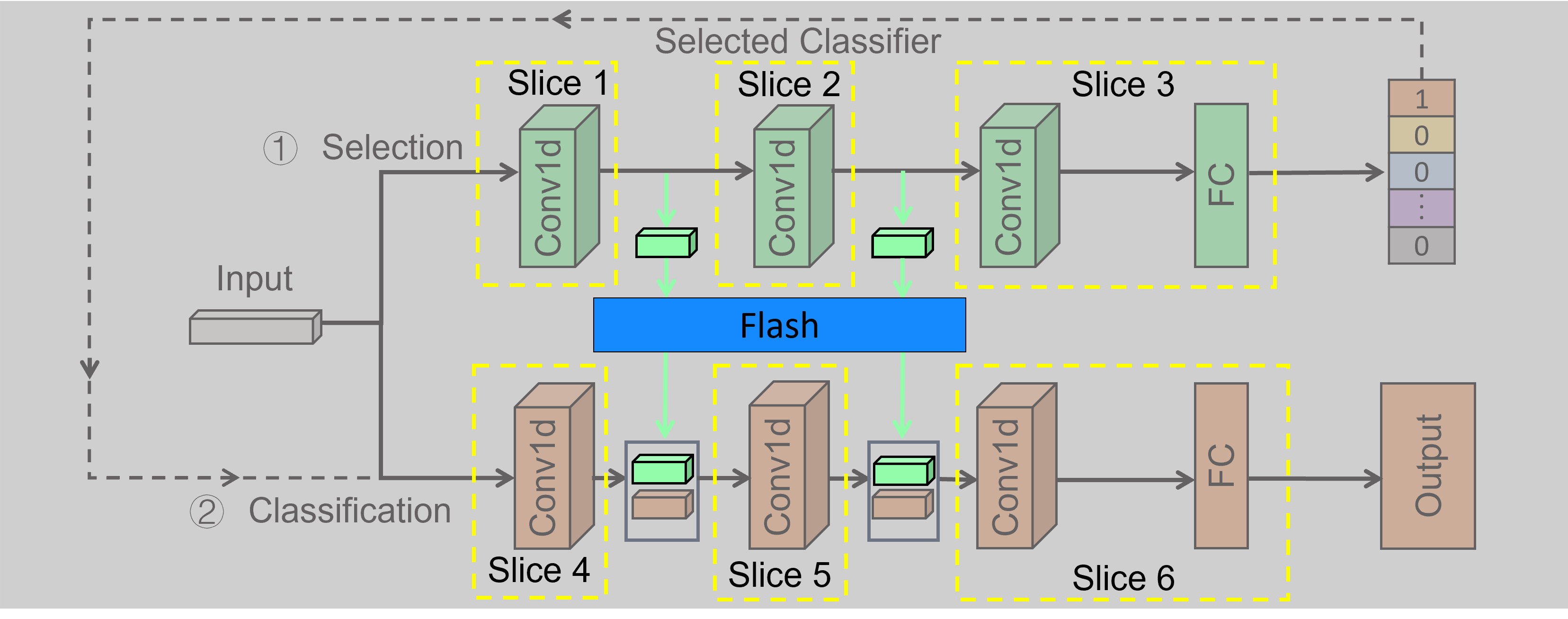}
  \vspace{0mm}
  \caption{Illustration of network slicing and the intermediate data flow.}
  \label{fig: network slicing}
   \vspace{-1mm}
\end{figure}

\section{Evaluation}

Next, we evaluate the performance of \sysname on different time series datasets and compare its classification accuracy with various baselines. We also measure the real system performance including latency, memory, and energy consumption.  

\subsection{Datasets}
We selected three public time series datasets that are expected to be executed on embedded devices in practice, covering typical pervasive sensing applications including human activity recognition (HAR), speech-based keyword spotting, and EEG-based emotion recognition. Each dataset is split into 80\% training and 20\% testing during the evaluation. 


\textbf{UniMiB-SHAR}~\cite{unimib}: A human activity recognition dataset collected with a 3-axis accelerometer on the smartphone. It consists of 17 classes (including 9 types of daily living activities and 8 types of falls) performed by 30 subjects. The total number of samples is 11,771, where each sample is a 3-second measurement sampled at 50~Hz.

\textbf{Speech Commands} ~\cite{speechcommand}: An audio dataset of spoken words collected by Google, which is designed for keyword spotting systems such as voice assistants. It contains 85,511 one-second long utterances of 30 short words spoken by thousands of different people. 
Given the high sampling rate of audio signals, we follow the audio processing convention to extract 26-channel MFCC features from raw audio files with 25~ms window size and 10~ms shift as the model input~\cite{poirier2019voice}. 

\textbf{DEAP}~\cite{deap}: An emotion recognition dataset that contains 32-channel electroencephalogram (EEG) signals of 32 volunteers when they are watching 40 one-minute-long excerpts of music videos. Following prior works \cite{saha2022automatic}, we downsample the dataset from 512~Hz to 128~Hz, and chop up the 32-channel EEG signals into 2-second frames with 1.875-second overlapping, resulting in a total of 148,480 samples. 
The original arousal-valence ratings
are converted to four emotion states based on the arousal-valence model \cite{houssein2022human}. 

\textit{Network Configuration: }In the following experiments, the strong model used for feature extraction during training data splitting adopts the same architecture as shown in \Cref{fig:architecture of strong model}, which achieves more than 95\% on all the three datasets. Since the difficulty of the three datasets varies, the default kernels of convolution layers in the selector and classifiers are [8,8,4,8,8,4] for UniMiB-SHAR, and [12,12,6,12,12,6] for both Speech Commands and DEAP. The default number of classifiers is set as 6 for all the datasets. Unless otherwise specified, all the results presented in the evaluation are based on the default settings.



\subsection{Baselines}
\label{sec: baselines}

To demonstrate the effectiveness of \sysname over existing techniques, we consider four baseline approaches listed below:

\textbf{Single Classifier (SigCla)}:
Essentially, by splitting the whole network into the selector and classifier, \sysname assigns the weights to two different tasks, where only half of the weights mainly focus on classification. Thus, our first baseline is to construct a network that has the same architecture (6 convolutional layers plus 2 FC layers) as \sysname (maintaining the same structure and number of parameters to ensure a fair comparison) while \textit{all} the weights are used for classification, referred to as the single classifier (SigCla). Then, the model will be trained from scratch on the full training dataset.


\textbf{Single Classifier with Knowledge Distillation (SigCla-KD)}:
In SigCla, although the whole network is utilized for classification, the performance could still be low as weak models have insufficient capacity to learn a concise knowledge representation from the data \cite{gou2021knowledge}. Knowledge distillation (KD), which transfers knowledge from a high-capacity model to a low-capacity model without loss of validity, has been considered state-of-the-art in training a weak model \cite{soro2021tinyml}. Thus, we treat the SigCla architecture trained with KD as our second baseline, referred to as SigCla-KD. Specifically, we apply a most recent KD framework \cite{huang2022knowledge} to the single classifier. The teacher model is the same as the strong model used for training data splitting in \sysname.


\textbf{Ensemble}: Model ensemble is another popular technique to improve classification performance using multiple weak models. Unlike \sysname which selects \textit{only} one weak model for classification, the ensemble combines \textit{all} the weak models for prediction. With the same computation overhead,  \sysname can be considered a two-model ensemble approach where the size of the weak models is equivalent to the selector (or classifier). Thus, to ensure fair comparison regarding computation during inference, we train two weak classifiers (with the same architecture as in \sysname) and combine them using the ensemble approach\cite{huang2017snapshot} for final classification. 


\textbf{Adapted Mixture of Experts (Ada-MoE)}:
Mixture of experts is a technique widely adopted in large models such as transformers to improve performance by having diverse experts while reducing the computation by activating part of the experts on a per-sample basis \cite{yuksel2012twenty}. The concept of MoE (i.e., routing the layer input to different experts via a small gating network) is similar to \sysname. 
Thus, to compare \sysname with MoE, we regard the selector as the gate and classifiers as experts, and train all of them together using a single cross-entropy loss (i.e., without the three core techniques of \sysname as presented in Section~\ref{sec: system_design}), referred to as Ada-MoE.

\subsection{Classification Performance}
\label{sec:eva_baseline}
We first compare the classification performance of \sysname with baselines and then evaluate the impact of different network configurations. All the results are averaged over 5 repeated experiments, each with random initialization.
\subsubsection{Comparison with baselines}
\begin{figure*}[t]
\centering
  \includegraphics[width = 0.8\textwidth]{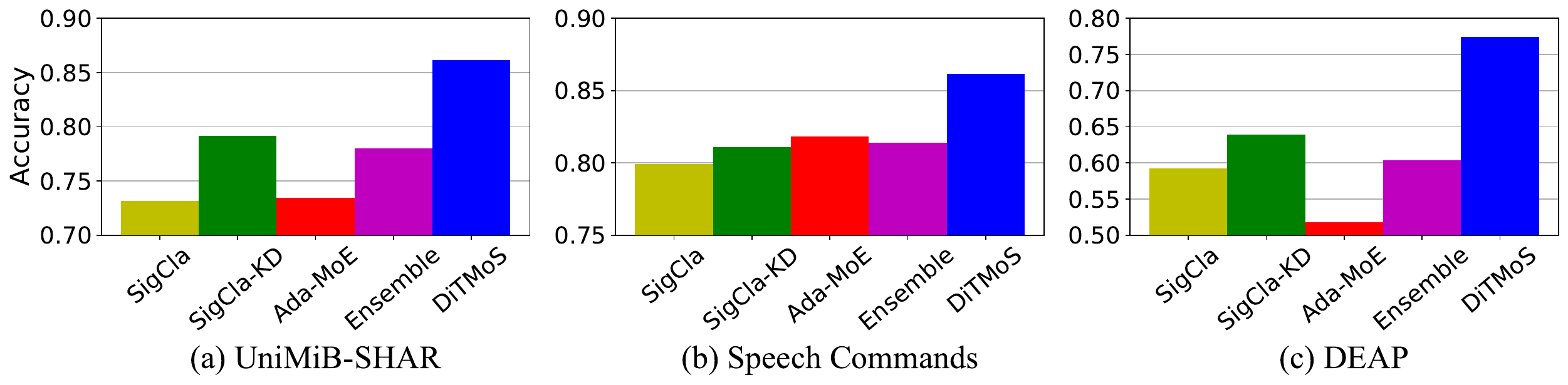}
  \vspace{0mm}
  \caption{Comparison of \sysname with different baselines. }
  \label{fig: overall performance}
\vspace{-2mm}
\end{figure*}

\begin{figure*}[t]
\centering
  \includegraphics[width = 0.8\textwidth]{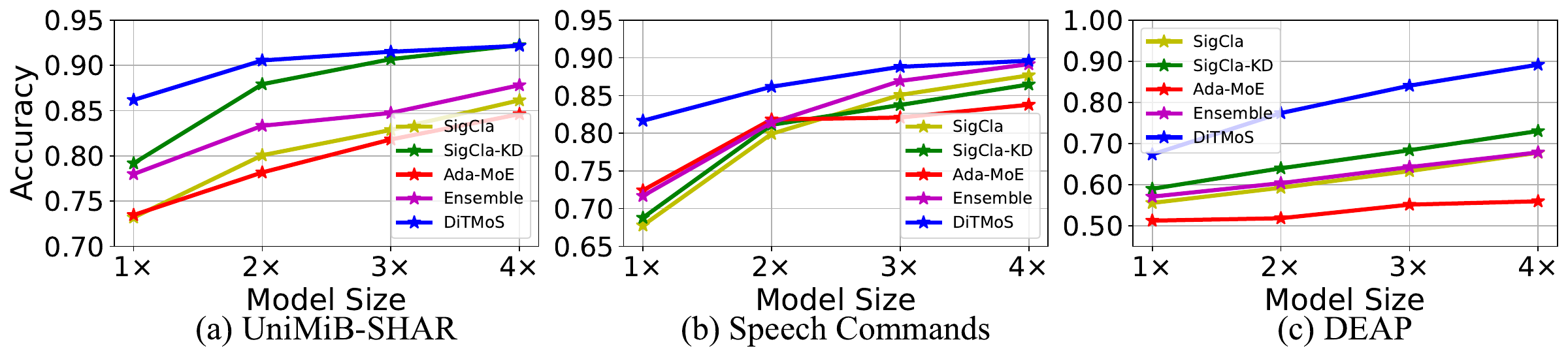}
  \vspace{0mm}
   \caption{Impact of model size on the three datasets. }
  \label{fig: model size}
   \vspace{-6mm}
\end{figure*}

\Cref{fig: overall performance} shows the classification accuracy of \sysname and other baselines on the three datasets, respectively. We can see that \sysname outperforms all the baselines with a large margin on all datasets. In detail, \sysname achieves 86.2\%, 86.2\%, and 77.4\% on UniMiB, Speech Commands, and DEAP, which are 7.0\%, 4.3\%, and 13.4\% higher than the best baseline respectively. In addition, the baseline approaches show large variations in different datasets. For example, Ada-MoE performs better than SigCla-KD on speech commands but performs very poorly on the other two datasets, as it is very sensitive to the datasets. The same finding is also presented in \cite{chen2022towards}, where the accuracy of MoE is even worse than a single model on the CIFAR-10 dataset without applying rotation.




\subsubsection{Impact of model size}


\begin{figure*}[t]
\centering
  \includegraphics[width = 0.8\textwidth]{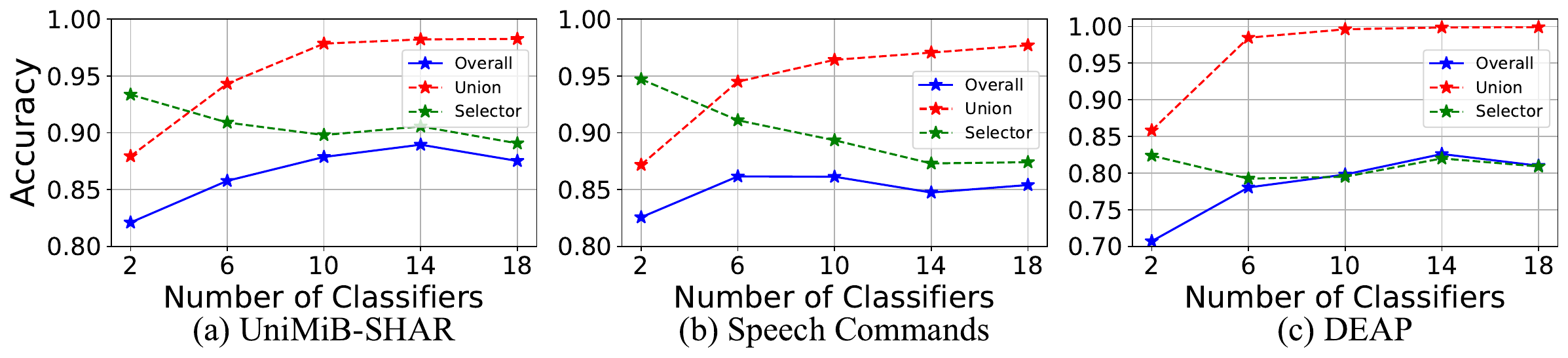}
  \vspace{0mm}
  \caption{Impact of the number of classifiers on the three datasets.}
  \label{fig: num classfiers}
   \vspace{-2mm}
\end{figure*}

Existing MCUs are equipped with memories of varying sizes, ranging from $<$16 KB for ultra-low-power MCUs, 16-256 KB for mainstream MCUs, to 256-512 KB for high-performance MCUs~\cite{stm32mainstream}. This allows application developers to design models with varying capacities to fit the accuracy and computation requirements flexibly. To evaluate how \sysname adapts to different MCU configurations, we design multiple networks with different sizes (from default model size to 4$\times$ of the default model size for each dataset) by changing the width of CNN and FC layer while keeping the same \textit{3 CNN+1 FC} architecture for the selector and classifiers. For example, for the UniMiB-SHAR dataset with CNN kernel size and FC size of [8,8,4,64], we construct a 2$\times$, 3$\times$, and 4$\times$ model by setting the layer size to [12,12,6,72], [16,16,8,112], and [20,20,10,144], respectively. The same procedure is applied to the other two datasets. 


\Cref{fig: model size} plots the accuracy change of different approaches over different model sizes on the three datasets. We can observe that (1) with the increase of the model size, all the methods achieve higher classification accuracy as the model is able to learn more accurate and fine-grained feature representations; (2) \sysname consistently outperforms other baselines across different model sizes, demonstrating the superior performance of \sysname on different MCU and application configurations; (3) on the UniMiB-SHAR and Speech Commands datasets, \sysname shows a higher improvement for smaller models. This is because our approach combats the weak capacity of smaller models using the diversity of multiple models.
Although Ada-MoE and ensemble also leverage model diversity, \sysname produces more diverse classifiers with the proposed training data splitting and the $CE_{overlap}$ loss to regularize the knowledge overlap between models. Concretely, the overlap of correctly classified samples among different models in the default setting for \sysname is 45\% lower than the value for Ada-MoE.

\subsubsection{Impact of number of classifiers}

Another design choice of \sysname is the number of classifiers, which can be adjusted by splitting the training dataset into different number of subsets, changing the output of the selector accordingly, and connecting the same number of classifiers to the selector. Intuitively, having more classifiers can create broader diversity thereby enhancing the union accuracy (i.e., the total number of samples that can be correctly classified by the classifiers) of the classifiers. However, this introduces more difficulty for the selector (i.e., a decrease of selector accuracy) as there are more classes. In addition, although the computation and memory cost per sample remain the same, more classifiers require larger Flash to store their weights. 

To explore the impact of the number of classifiers, we adjusted it from 2 to 18 on the three datasets. \Cref{fig: num classfiers} shows the corresponding union accuracy, selector accuracy, and overall accuracy for each dataset. We can observe that (1) there is indeed a trade-off between the union accuracy and selector accuracy, which leads to an optimal number of classifiers in terms of the overall accuracy; (2) however, this optimal number is different for different datasets as the difficulty of the three tasks varies. The optimal number might be large for specific datasets, leading to notable Flash overhead when storing the classifiers. Thus, application designers can flexibly adjust the number of classifiers based on the overall accuracy curve and the available Flash resource. The results in \Cref{fig: num classfiers} also indicate that even under the non-optimal settings, our framework can achieve higher accuracy than baselines; (3) since the optimal numbers for some datasets are larger than our default setting (i.e., 6), the results indicate that \sysname can achieve even more accuracy improvement than we presented in Section \ref{sec:eva_baseline}-1), compared to the baselines.

\begin{table}[t]
\caption{Ablation study of DiTMoS with default setting.}
\vspace{0mm}
\label{tab:ablation study}
\resizebox{\linewidth}{!}{ 
\begin{tabular}{c|c|c|c}
\toprule
Ablation Study               & UniMiB-SHAR & Speech Commands & DEAP    \\ \hline

Random Splitting            & 84.9\%     & 84.3\%         & 75.8\% \\ \hline

w/o Adversarial Training & 72.6\%     & 81.8\%         & 56.3\% \\ \hline

w/o Feature Aggregation        & 83.5\%     & 86.0\%         & 76.3\% \\ \hline

\textbf{DiTMoS}                       & \textbf{86.2}\%     & \textbf{86.2}\%         & \textbf{77.4}\% \\ \bottomrule
\end{tabular}}
 \vspace{-3mm}
\end{table}

\subsubsection{Ablation Study}
Next, we present the ablation study that assesses the effectiveness of different components in \sysname. Specifically, for training data splitting, instead of using a strong model to extract more representative features, we randomly split the dataset into multiple subsets without K-Means. For adversarial training, we just train the selector and classifiers together in a synchronous manner, similar to MoE. For heterogeneous feature aggregation, the intermediate embeddings from  selector are excluded for classifier training. 

\Cref{tab:ablation study} reports the performance of the three datasets. 
First, we can observe that data splitting with a strong model contributes 2\% improvement compared to random splitting
as it not only increases the similarity of samples within each subset but also enhances the diversity of different subsets. Second, adversarial training is the most effective technique in \sysname and dramatically lifts the recognition accuracy. This implies that compared to synchronous training, adversarial training can enable more strongly coupled selector and classifiers which work coordinately to make a prediction.
Third, heterogeneous feature aggregation can also enhance the overall accuracy, although the improvement is smaller on some datasets. The main reason is that the default model is very small so the features from the selector contain little knowledge that can be helpful in classifiers. We then change the model size to be 3$\times$ and 4$\times$ larger, the accuracy without feature aggregation drops by 3.2\% and 3.5\% respectively on the DEAP dataset, revealing the potential of feature aggregation.

\subsection{System Performance}
\label{sec: system_performance}

Finally, we evaluate the system performance of \sysname based on the UniMiB-SHAR dataset. Specifically, we utilize the memory analyzer in STM32CubeIDE to measure the peak memory usage, and a USB power meter to measure the power consumption. Flash usage is basically the size of the models. \Cref{tab:system performance} presents the results, which are based on multiple runs of inference and the latency is averaged over 1000 runs.\footnote{Note that the accuracy on MCU is the same as that measured on the server because we did not apply quantization (i.e., the precision of weights and activations is the same. Thus, we only report the system overheads in this subsection.}


\subsubsection{Memory \& Flash Usage}
As presented in Section \ref{sec: implemtation}, with the layer-by-layer loading and execution strategy, the peak memory usage is constrained by the layer that has the largest input and output sizes. Thus, all four baselines have the same peak memory. For our approach, heterogeneous feature aggregation requires reserving the intermediate activations of the selector until they are aggregated in the classifier, which incurs additional memory usage depending on the size of the network (2.4~KB for the default network). However, with the proposed network slicing technique, the additional memory usage can be eliminated (the extra 0.1~KB usage is due to the increased channels when reusing the features). In terms of Flash, although Ada-MoE and \sysname require more usage depending on the number of experts/classifiers, it is affordable as MCU usually equips much larger Flash than memory. 

\begin{table}[t]
\caption{System performance on UniMiB-SHAR.}
\vspace{0mm}
\label{tab:system performance}
\resizebox{\columnwidth}{!}{%
\begin{tabular}{c|c|c|c|c}
\toprule
Approach & \begin{tabular}[c]{@{}c@{}}Memory\\  Usage(KB)\end{tabular} & \begin{tabular}[c]{@{}c@{}}Flash\\ Usage(KB)\end{tabular} & \begin{tabular}[c]{@{}c@{}}Latency\\ (ms)\end{tabular} & \begin{tabular}[c]{@{}c@{}}Energy\\ (mJ)\end{tabular} \\ \hline
SigCla & 6.1 & 63.6 & 11.9 & 3.9 \\ \hline
SigCla-KD & 6.1 & 63.6 & 11.9 & 3.9 \\ \hline
Ada-MoE & 6.1 & 168.2 & 10.4 & 3.4 \\ \hline
Ensemble & 6.1 & 51.4 & 10.4 & 3.4 \\ \hline
\textbf{DiTMoS w/o Slicing} & \textbf{8.5} & \textbf{166.9} & \textbf{10.9} & \textbf{3.6} \\ \hline
\textbf{DiTMoS} & \textbf{6.2} & \textbf{166.9} & \textbf{12.5} & \textbf{4.1} \\ \bottomrule
\end{tabular}%
}
\vspace{-2mm}
\end{table}



\subsubsection{Energy and Latency}


As mentioned in Section \ref{sec: baselines}, the sizes of the baseline models are devised to ensure a fair comparison in terms of computation cost during inference. Thus, the latency and energy consumption for all the baselines are almost the same. For \sysname, heterogeneous feature aggregation increases the latency in two aspects: (1) storing and loading the intermediate activations using Flash requires 1.6~ms and (2) computing the aggregated activations incurs 0.5~ms overhead. The first part is introduced by network slicing, which trade-offs between memory usage and latency. With feature aggregation, the second part is an unavoidable but fortunately minor overhead. In practice, for latency-sensitive applications, developers can forgo feature aggregation in order to reduce latency.  \Cref{fig: overall performance} and \Cref{tab:ablation study} indicate that DiTMoS still achieves better performance than other baselines without feature aggregation.


\section{Discussion}

\textit{Orthogonal to model compression}:
To enable DNN execution on MCUs,
existing efforts mainly revolve around compressing a large and accurate model into a smaller one, often involving a modest compromise in accuracy. Techniques such as pruning \cite{liberis2023differentiable} and quantization \cite{daghero2022human} exemplify this approach. However, we offer a fresh perspective by constructing a cohort of weak yet diverse models and selecting the most suitable model for classification. Consequently, our approach operates independently and orthogonally of traditional model compression techniques, allowing for their combined use to enhance performance. For instance, by applying pruning and quantization, \sysname can construct larger models (thus with larger capacity) to improve the overall accuracy while maintaining the same memory usage and computation overhead.


\textit{Flexible design of \sysname}: The core of \sysname is the concept of selector-classifiers architecture, which leverages model diversity (translates to a high union accuracy) and selector-classifiers coupling to improve the overall performance. In this work, we consider a balanced selector-classifiers network configuration to demonstrate the concept, where the selector and classifiers equally share the whole network capacity by having the same architecture. However, the capability of each sub-network can be adjusted flexibly to accommodate the characteristics of the task for better performance.
Moreover, instead of having the same structure for all classifiers, each classifier can be designed with different structures, kernel sizes, and the number of kernels, which have been proven to be helpful in increasing the diversity of models \cite{tu2022nas,nguyen2020widedeep}.

\section{Related Work}

\textit{Embedded machine learning}: it has been an active research field in recent years and numerous approaches have been proposed.
We summarize these works in three threads: (1) frameworks to support end-to-end DNN inference on MCU: TensorFlow Lite for Microcontrollers (TFLM)~\cite{david2021tensorflow} is an interpreter-based tool-chain which converts Tensorflow Lite models to C-code based on an operator library.
MCUNet~\cite{lin2020mcunet} is a system-algorithm co-design framework consisting of a TinyNAS module to search for the best model architecture 
and a TinyEngine module to optimize runtime memory usage; 
(2) model compression: Liberis et al.~\cite{liberis2023differentiable} proposed a differentiable pruning structure that iteratively removes unnecessary weights based on specific resource budgets and channel importance, while BC-Net~\cite{mocerino2021bcnet} optimize computation by combining binarization and quantization;
and (3) memory management: to reduce the peak memory usage, researchers proposed to decompose a full convolutional layer into small patches~\cite{lin2021memory} or only compute partial regions across multiple layers~\cite{liberis2022pex}, or reordering operations in multi-branch networks~\cite{liberis2019reorder}. Unlike them, \sysname is designed based on the concept of model selection, which has not been explored in the context of embedded machine learning.

\textit{Ensemble \& MoE}: \sysname leverages the model diversity to gain performance improvement so our work is also related to ensemble learning \cite{ani2017iot} and the mixture of experts (MoE) \cite{gong2022real}. Specifically, compared to ensemble learning that combines all the weaker models for each inference, \sysname only selects one classifier, thereby significantly reducing the computation overhead. The pipeline of \sysname is similar to MoE, while they differ in that (1) the expert in MoE is just part of the network (e.g., a block in a layer) instead of a standalone model; (2) MoE usually selects more than one expert and combines them for prediction; and (3) the gate and experts in MoE are trained synchronously. Notably, Taylor et al. \cite{taylor2018adaptive} proposed a KNN-based model selection scheme to select a model (either small, medium, or large) for each input. However, the aim is to reduce the computation cost and the three models just differ in capacity instead of diversity.

\section{Conclusion}

This paper presents \sysname, a memory-efficient selector-classifier framework that enables accurate inference of time series classification on MCUs. 
\sysname consists of 3 major components: training data splitting to enlarge the diversity of classifiers, adversarial training to coordinate the selector and classifier, and heterogeneous feature aggregation to enhance the classifier capability. We further proposed a network slicing technique to optimize memory usage. The experiment result shows that \sysname significantly outperforms baselines while incurring marginal latency and memory overhead. 


\bibliographystyle{IEEEtran}
\bibliography{Reference}

\begin{thebibliography}{10}
\providecommand{\url}[1]{#1}
\csname url@samestyle\endcsname
\providecommand{\newblock}{\relax}
\providecommand{\bibinfo}[2]{#2}
\providecommand{\BIBentrySTDinterwordspacing}{\spaceskip=0pt\relax}
\providecommand{\BIBentryALTinterwordstretchfactor}{4}
\providecommand{\BIBentryALTinterwordspacing}{\spaceskip=\fontdimen2\font plus
\BIBentryALTinterwordstretchfactor\fontdimen3\font minus
  \fontdimen4\font\relax}
\providecommand{\BIBforeignlanguage}[2]{{%
\expandafter\ifx\csname l@#1\endcsname\relax
\typeout{** WARNING: IEEEtran.bst: No hyphenation pattern has been}%
\typeout{** loaded for the language `#1'. Using the pattern for}%
\typeout{** the default language instead.}%
\else
\language=\csname l@#1\endcsname
\fi
#2}}
\providecommand{\BIBdecl}{\relax}
\BIBdecl

\bibitem{ullo2020iotenvironment}
S.~L. Ullo and G.~R. Sinha, ``Advances in smart environment monitoring systems
  using iot and sensors,'' \emph{Sensors}, vol.~20, no.~11, p. 3113, 2020.

\bibitem{khalid2022iotasset}
R.~Khalid and W.~Ejaz, ``Internet of things-based on-demand rental asset
  tracking and monitoring system,'' in \emph{2022 5th International Conference
  on Information and Computer Technologies (ICICT)}.\hskip 1em plus 0.5em minus
  0.4em\relax IEEE, 2022, pp. 84--89.

\bibitem{anikwe2022iothealth}
C.~V. Anikwe, H.~F. Nweke, A.~C. Ikegwu, C.~A. Egwuonwu, F.~U. Onu, U.~R. Alo,
  and Y.~W. Teh, ``Mobile and wearable sensors for data-driven health
  monitoring system: State-of-the-art and future prospect,'' \emph{Expert
  Systems with Applications}, vol. 202, p. 117362, 2022.

\bibitem{zhang2022deepsensingreview}
S.~Zhang, Y.~Li, S.~Zhang, F.~Shahabi, S.~Xia, Y.~Deng, and N.~Alshurafa,
  ``Deep learning in human activity recognition with wearable sensors: A review
  on advances,'' \emph{Sensors}, vol.~22, no.~4, p. 1476, 2022.

\bibitem{liu2022swin}
Z.~Liu, H.~Hu, Y.~Lin, Z.~Yao, Z.~Xie, Y.~Wei, J.~Ning, Y.~Cao, Z.~Zhang,
  L.~Dong \emph{et~al.}, ``Swin transformer v2: Scaling up capacity and
  resolution,'' in \emph{Proceedings of the IEEE/CVF conference on computer
  vision and pattern recognition}, 2022, pp. 12\,009--12\,019.

\bibitem{lin2011large}
Y.~Lin, F.~Lv, S.~Zhu, M.~Yang, T.~Cour, K.~Yu, L.~Cao, and T.~Huang,
  ``Large-scale image classification: Fast feature extraction and svm
  training,'' in \emph{CVPR 2011}.\hskip 1em plus 0.5em minus 0.4em\relax IEEE,
  2011, pp. 1689--1696.

\bibitem{stm32mainstream}
``Stm32 mainstream microcontrollers,''
  \url{https://www.st.com/en/microcontrollers-microprocessors/stm32-mainstream-mcus.html}.

\bibitem{banbury2021micronets}
C.~Banbury, C.~Zhou, I.~Fedorov, R.~Matas, U.~Thakker, D.~Gope,
  V.~Janapa~Reddi, M.~Mattina, and P.~Whatmough, ``Micronets: Neural network
  architectures for deploying tinyml applications on commodity
  microcontrollers,'' \emph{Proceedings of Machine Learning and Systems},
  vol.~3, pp. 517--532, 2021.

\bibitem{garbay2021cost}
T.~Garbay, P.~Dobias, W.~Dron, P.~Lusich, I.~Khalis, A.~Pinna, K.~Hachicha, and
  B.~Granado, ``Cnn inference costs estimation on microcontrollers: the est
  primitive-based model,'' in \emph{2021 28th IEEE International Conference on
  Electronics, Circuits, and Systems (ICECS)}.\hskip 1em plus 0.5em minus
  0.4em\relax IEEE, 2021, pp. 1--5.

\bibitem{liberis2023differentiable}
E.~Liberis and N.~D. Lane, ``Differentiable neural network pruning to enable
  smart applications on microcontrollers,'' \emph{Proceedings of the ACM on
  Interactive, Mobile, Wearable and Ubiquitous Technologies}, vol.~6, no.~4,
  pp. 1--19, 2023.

\bibitem{daghero2022human}
F.~Daghero, A.~Burrello, C.~Xie, M.~Castellano, L.~Gandolfi, A.~Calimera,
  E.~Macii, M.~Poncino, and D.~J. Pagliari, ``Human activity recognition on
  microcontrollers with quantized and adaptive deep neural networks,''
  \emph{ACM Transactions on Embedded Computing Systems (TECS)}, vol.~21, no.~4,
  pp. 1--28, 2022.

\bibitem{cerutti2020compact}
G.~Cerutti, R.~Prasad, A.~Brutti, and E.~Farella, ``Compact recurrent neural
  networks for acoustic event detection on low-energy low-complexity
  platforms,'' \emph{IEEE Journal of Selected Topics in Signal Processing},
  vol.~14, no.~4, pp. 654--664, 2020.

\bibitem{brutti2022optimizing}
A.~Brutti, F.~Paissan, A.~Ancilotto, and E.~Farella, ``Optimizing phinet
  architectures for the detection of urban sounds on low-end devices,'' in
  \emph{2022 30th European Signal Processing Conference (EUSIPCO)}.\hskip 1em
  plus 0.5em minus 0.4em\relax IEEE, 2022, pp. 1121--1125.

\bibitem{yang2021taxonomizing}
Y.~Yang, L.~Hodgkinson, R.~Theisen, J.~Zou, J.~E. Gonzalez, K.~Ramchandran, and
  M.~W. Mahoney, ``Taxonomizing local versus global structure in neural network
  loss landscapes,'' \emph{Advances in Neural Information Processing Systems},
  vol.~34, pp. 18\,722--18\,733, 2021.

\bibitem{huang2022deep}
W.~Huang, L.~Zhang, S.~Wang, H.~Wu, and A.~Song, ``Deep ensemble learning for
  human activity recognition using wearable sensors via filter activation,''
  \emph{ACM Transactions on Embedded Computing Systems}, vol.~22, no.~1, pp.
  1--23, 2022.

\bibitem{unimib}
D.~Micucci, M.~Mobilio, and P.~Napoletano, ``Unimib shar: A dataset for human
  activity recognition using acceleration data from smartphones,''
  \emph{Applied Sciences}, vol.~7, no.~10, 2017.

\bibitem{kornblith2019similarity}
S.~Kornblith, M.~Norouzi, H.~Lee, and G.~Hinton, ``Similarity of neural network
  representations revisited,'' in \emph{International conference on machine
  learning}.\hskip 1em plus 0.5em minus 0.4em\relax PMLR, 2019, pp. 3519--3529.

\bibitem{macqueen1967some}
J.~MacQueen \emph{et~al.}, ``Some methods for classification and analysis of
  multivariate observations,'' in \emph{Proceedings of the fifth Berkeley
  symposium on mathematical statistics and probability}, vol.~1, no.~14.\hskip
  1em plus 0.5em minus 0.4em\relax Oakland, CA, USA, 1967, pp. 281--297.

\bibitem{stm32f767zi}
``Stm32 nucleo-f767zi board,''
  \url{https://www.st.com/en/microcontrollers-microprocessors/stm32f767zi.html}.

\bibitem{liberis2022pex}
E.~Liberis and N.~D. Lane, ``Pex: Memory-efficient microcontroller deep
  learning through partial execution,'' \emph{arXiv preprint arXiv:2211.17246},
  2022.

\bibitem{onnx}
``Open neural network exchange(onnx),'' \url{https://onnx.ai/}.

\bibitem{liberis2019reorder}
E.~Liberis and N.~D. Lane, ``Neural networks on microcontrollers: saving memory
  at inference via operator reordering,'' \emph{arXiv preprint
  arXiv:1910.05110}, 2019.

\bibitem{speechcommand}
P.~Warden, ``Speech commands: A dataset for limited-vocabulary speech
  recognition,'' \emph{arXiv preprint arXiv:1804.03209}, 2018.

\bibitem{poirier2019voice}
S.~Poirier, F.~Routhier, and A.~Campeau-Lecours, ``Voice control interface
  prototype for assistive robots for people living with upper limb
  disabilities,'' in \emph{2019 IEEE 16th International Conference on
  Rehabilitation Robotics (ICORR)}.\hskip 1em plus 0.5em minus 0.4em\relax
  IEEE, 2019, pp. 46--52.

\bibitem{deap}
S.~Koelstra, C.~Muhl, M.~Soleymani, J.-S. Lee, A.~Yazdani, T.~Ebrahimi, T.~Pun,
  A.~Nijholt, and I.~Patras, ``Deap: A database for emotion analysis ;using
  physiological signals,'' \emph{IEEE Transactions on Affective Computing},
  vol.~3, no.~1, pp. 18--31, 2012.

\bibitem{saha2022automatic}
O.~Saha, M.~S. Mahmud, S.~A. Fattah, and M.~Saquib, ``Automatic emotion
  recognition from multi-band eeg data based on a deep learning scheme with
  effective channel attention,'' \emph{IEEE Access}, vol.~11, pp. 2342--2350,
  2022.

\bibitem{houssein2022human}
E.~H. Houssein, A.~Hammad, and A.~A. Ali, ``Human emotion recognition from
  eeg-based brain--computer interface using machine learning: a comprehensive
  review,'' \emph{Neural Computing and Applications}, vol.~34, no.~15, pp.
  12\,527--12\,557, 2022.

\bibitem{gou2021knowledge}
J.~Gou, B.~Yu, S.~J. Maybank, and D.~Tao, ``Knowledge distillation: A survey,''
  \emph{International Journal of Computer Vision}, vol. 129, pp. 1789--1819,
  2021.

\bibitem{soro2021tinyml}
S.~Soro, ``Tinyml for ubiquitous edge ai,'' \emph{arXiv preprint
  arXiv:2102.01255}, 2021.

\bibitem{huang2022knowledge}
T.~Huang, S.~You, F.~Wang, C.~Qian, and C.~Xu, ``Knowledge distillation from a
  stronger teacher,'' \emph{Advances in Neural Information Processing Systems},
  vol.~35, pp. 33\,716--33\,727, 2022.

\bibitem{huang2017snapshot}
G.~Huang, Y.~Li, G.~Pleiss, Z.~Liu, J.~E. Hopcroft, and K.~Q. Weinberger,
  ``Snapshot ensembles: Train 1, get m for free,'' \emph{arXiv preprint
  arXiv:1704.00109}, 2017.

\bibitem{yuksel2012twenty}
S.~E. Yuksel, J.~N. Wilson, and P.~D. Gader, ``Twenty years of mixture of
  experts,'' \emph{IEEE transactions on neural networks and learning systems},
  vol.~23, no.~8, pp. 1177--1193, 2012.

\bibitem{chen2022towards}
Z.~Chen, Y.~Deng, Y.~Wu, Q.~Gu, and Y.~Li, ``Towards understanding the
  mixture-of-experts layer in deep learning,'' in \emph{Advances in Neural
  Information Processing Systems}, S.~Koyejo, S.~Mohamed, A.~Agarwal,
  D.~Belgrave, K.~Cho, and A.~Oh, Eds., vol.~35, 2022, pp. 23\,049--23\,062.

\bibitem{tu2022nas}
R.~Tu, N.~Roberts, M.~Khodak, J.~Shen, F.~Sala, and A.~Talwalkar,
  ``Nas-bench-360: Benchmarking neural architecture search on diverse tasks,''
  \emph{Advances in Neural Information Processing Systems}, vol.~35, pp.
  12\,380--12\,394, 2022.

\bibitem{nguyen2020widedeep}
T.~Nguyen, M.~Raghu, and S.~Kornblith, ``Do wide and deep networks learn the
  same things? uncovering how neural network representations vary with width
  and depth,'' \emph{arXiv preprint arXiv:2010.15327}, 2020.

\bibitem{david2021tensorflow}
R.~David, J.~Duke, A.~Jain, V.~Janapa~Reddi, N.~Jeffries, J.~Li, N.~Kreeger,
  I.~Nappier, M.~Natraj, T.~Wang \emph{et~al.}, ``Tensorflow lite micro:
  Embedded machine learning for tinyml systems,'' \emph{Proceedings of Machine
  Learning and Systems}, vol.~3, pp. 800--811, 2021.

\bibitem{lin2020mcunet}
J.~Lin, W.-M. Chen, Y.~Lin, C.~Gan, S.~Han \emph{et~al.}, ``Mcunet: Tiny deep
  learning on iot devices,'' \emph{Advances in Neural Information Processing
  Systems}, vol.~33, pp. 11\,711--11\,722, 2020.

\bibitem{mocerino2021bcnet}
L.~Mocerino and A.~Calimera, ``Fast and accurate inference on microcontrollers
  with boosted cooperative convolutional neural networks (bc-net),'' \emph{IEEE
  Transactions on Circuits and Systems I: Regular Papers}, vol.~68, no.~1, pp.
  77--88, 2021.

\bibitem{lin2021memory}
J.~Lin, W.-M. Chen, H.~Cai, C.~Gan, and S.~Han, ``Memory-efficient patch-based
  inference for tiny deep learning,'' \emph{Advances in Neural Information
  Processing Systems}, vol.~34, pp. 2346--2358, 2021.

\bibitem{ani2017iot}
R.~Ani, S.~Krishna, N.~Anju, M.~S. Aslam, and O.~Deepa, ``Iot based patient
  monitoring and diagnostic prediction tool using ensemble classifier,'' in
  \emph{2017 International conference on advances in computing, communications
  and informatics (ICACCI)}.\hskip 1em plus 0.5em minus 0.4em\relax IEEE, 2017,
  pp. 1588--1593.

\bibitem{gong2022real}
X.~Gong, Q.~Feng, Y.~Zhang, J.~Qin, W.~Ding, B.~Li, P.~Jiang, and K.~Gai,
  ``Real-time short video recommendation on mobile devices,'' in
  \emph{Proceedings of the 31st ACM International Conference on Information \&
  Knowledge Management}, 2022, pp. 3103--3112.

\bibitem{taylor2018adaptive}
B.~Taylor, V.~S. Marco, W.~Wolff, Y.~Elkhatib, and Z.~Wang, ``Adaptive deep
  learning model selection on embedded systems,'' \emph{ACM SIGPLAN Notices},
  vol.~53, no.~6, pp. 31--43, 2018.

\end{thebibliography}
\end{document}